\documentclass[11pt]{article}

\PassOptionsToPackage{dvipsnames,table,xcdraw}{xcolor}

\usepackage[final]{acl}

\usepackage{times}
\usepackage[utf8]{inputenc}

\usepackage{makecell}
\usepackage{tabularx}

\usepackage{latexsym}

\usepackage[T1]{fontenc}

\usepackage{microtype}

\usepackage{inconsolata}

\usepackage{graphicx}
\usepackage[dvipsnames,table,xcdraw]{xcolor}
\usepackage{amsmath, amssymb}
\usepackage{tcolorbox}
\tcbuselibrary{listings,skins,breakable}

\usepackage{lipsum}               
\tcbuselibrary{breakable}
\usepackage{cuted}
\usepackage{fancyvrb}
\usepackage{listings}
\usepackage{booktabs}
\usepackage{multirow}
\usepackage{algorithm}
\usepackage{algorithmic}
\usepackage{geometry}
\usepackage{subcaption}
\usepackage{float}
\usepackage{rotating}
\usepackage{wrapfig}
\usepackage{pifont}
\usepackage{enumitem}

\tcbuselibrary{breakable, skins}
\usepackage{dblfloatfix}

\definecolor{skyblue}{RGB}{135,206,235}
\definecolor{lemonchiffon}{RGB}{255,250,205}
\definecolor{lightcoral}{RGB}{240,128,128}
\colorlet{oursrow}{SeaGreen!10}

\newcommand{\besti}[1]{\textbf{#1}}
\newcommand{\secondi}[1]{\underline{#1}}
\newcommand{\ms}[2]{#1{\scriptsize\,$\pm$\,#2}}

\definecolor{prompt}{RGB}{173, 216, 230}
\definecolor{prompt-frame}{RGB}{0, 51, 102}
\definecolor{prompt2}{RGB}{173, 216, 230}
\definecolor{prompt2-frame}{RGB}{0, 51, 102}
\definecolor{prompt3}{RGB}{173, 216, 230}
\definecolor{prompt3-frame}{RGB}{0, 51, 102}
\definecolor{prompt4}{RGB}{173, 216, 230}
\definecolor{prompt4-frame}{RGB}{0, 51, 102}
\definecolor{prompt5}{RGB}{173, 216, 230}
\definecolor{prompt5-frame}{RGB}{0, 51, 102}
\usepackage{fontawesome5}
\tcbset{
    prompt_func/.style={
        enhanced,
        breakable,
        colback=prompt,        
        colframe=prompt-frame,            
        fontupper=\normalsize,               
        boxrule=1pt,                       
        arc=4mm,                           
        left=1mm, right=1mm, top=1mm, bottom=1mm, 
        boxsep=4pt,                        
        before skip=10pt, after skip=10pt, 
        overlay={
            \node[anchor=north west, xshift=4pt, text=white] at (frame.north west) {\faDatabase};
        },
        title={~~~~~~~\textbf{Prompt used with GPT-4o to generate annotation for Element Grounding with functional setting}},
        coltitle=white,
        fonttitle=\bfseries\small
    }
}

\tcbset{
    full_content_generation/.style={
        enhanced,
        breakable,
        colback=prompt,        
        colframe=prompt-frame,            
        fontupper=\normalsize,               
        boxrule=1pt,                       
        arc=4mm,                           
        left=1mm, right=1mm, top=1mm, bottom=1mm, 
        boxsep=4pt,                        
        before skip=10pt, after skip=10pt, 
        title={\faDatabase~~~\textbf{Prompt: Full Content Generation}},
        coltitle=white,
        fonttitle=\bfseries\small
    }
}
\tcbset{
    filter_figures/.style={
        enhanced,
        breakable,
        colback=prompt,        
        colframe=prompt-frame,            
        fontupper=\normalsize,               
        boxrule=1pt,                       
        arc=4mm,                           
        left=1mm, right=1mm, top=1mm, bottom=1mm, 
        boxsep=4pt,                        
        before skip=10pt, after skip=10pt, 
        title={\faDatabase~~~\textbf{Prompt: Filter Figures}},
        coltitle=white,
        fonttitle=\bfseries\small
    }
}
\tcbset{
    section_generation/.style={
        enhanced,
        breakable,
        colback=prompt,        
        colframe=prompt-frame,            
        fontupper=\normalsize,               
        boxrule=1pt,                       
        arc=4mm,                           
        left=1mm, right=1mm, top=1mm, bottom=1mm, 
        boxsep=4pt,                        
        before skip=10pt, after skip=10pt, 
        title={\faDatabase~~~\textbf{Prompt: Section Generation}},
        coltitle=white,
        fonttitle=\bfseries\small
    }
}
\tcbset{
    html_generation/.style={
        enhanced,
        breakable,
        colback=prompt,        
        colframe=prompt-frame,            
        fontupper=\normalsize,               
        boxrule=1pt,                       
        arc=4mm,                           
        left=1mm, right=1mm, top=1mm, bottom=1mm, 
        boxsep=4pt,                        
        before skip=10pt, after skip=10pt, 
        title={\faDatabase~~~\textbf{Prompt: HTML Generation}},
        coltitle=white,
        fonttitle=\bfseries\small
    }
}
\tcbset{
    full_content_review/.style={
        enhanced,
        breakable,
        colback=prompt,        
        colframe=prompt-frame,            
        fontupper=\normalsize,               
        boxrule=1pt,                       
        arc=4mm,                           
        left=1mm, right=1mm, top=1mm, bottom=1mm, 
        boxsep=4pt,                        
        before skip=10pt, after skip=10pt, 
        title={\faDatabase~~~\textbf{Prompt: Full Content Review}},
        coltitle=white,
        fonttitle=\bfseries\small
    }
}
\tcbset{
    full_content_revise/.style={
        enhanced,
        breakable,
        colback=prompt,        
        colframe=prompt-frame,            
        fontupper=\normalsize,               
        boxrule=1pt,                       
        arc=4mm,                           
        left=1mm, right=1mm, top=1mm, bottom=1mm, 
        boxsep=4pt,                        
        before skip=10pt, after skip=10pt, 
        title={\faDatabase~~~\textbf{Prompt: Full Content Revise}},
        coltitle=white,
        fonttitle=\bfseries\small
    }
}
\tcbset{
    text_content_generation/.style={
        enhanced,
        breakable,
        colback=prompt,        
        colframe=prompt-frame,            
        fontupper=\normalsize,               
        boxrule=1pt,                       
        arc=4mm,                           
        left=1mm, right=1mm, top=1mm, bottom=1mm, 
        boxsep=4pt,                        
        before skip=10pt, after skip=10pt, 
        title={\faDatabase~~~\textbf{Prompt: Text Content Generation}},
        coltitle=white,
        fonttitle=\bfseries\small
    }
}
\tcbset{
    html_review/.style={
        enhanced,
        breakable,
        colback=prompt,        
        colframe=prompt-frame,            
        fontupper=\normalsize,               
        boxrule=1pt,                       
        arc=4mm,                           
        left=1mm, right=1mm, top=1mm, bottom=1mm, 
        boxsep=4pt,                        
        before skip=10pt, after skip=10pt, 
        title={\faDatabase~~~\textbf{Prompt: HTML Review}},
        coltitle=white,
        fonttitle=\bfseries\small
    }
}
\tcbset{
    html_revise/.style={
        enhanced,
        breakable,
        colback=prompt,        
        colframe=prompt-frame,            
        fontupper=\normalsize,               
        boxrule=1pt,                       
        arc=4mm,                           
        left=1mm, right=1mm, top=1mm, bottom=1mm, 
        boxsep=4pt,                        
        before skip=10pt, after skip=10pt, 
        title={\faDatabase~~~\textbf{Prompt: HTML Revise}},
        coltitle=white,
        fonttitle=\bfseries\small
    }
}
\tcbset{
    prompt_aethetic_element/.style={
        enhanced,
        breakable,
        colback=prompt,        
        colframe=prompt-frame,            
        fontupper=\normalsize,               
        boxrule=1pt,                       
        arc=4mm,                           
        left=1mm, right=1mm, top=1mm, bottom=1mm, 
        boxsep=4pt,                        
        before skip=10pt, after skip=10pt, 
        title={\faDatabase~~~\textbf{Prompt: Element Quality Judge}},
        coltitle=white,
        fonttitle=\bfseries\small
    }
}

\tcbset{
    prompt_aethetic_Layout/.style={
        enhanced,
        breakable,
        colback=prompt,        
        colframe=prompt-frame,            
        fontupper=\normalsize,               
        boxrule=1pt,                       
        arc=4mm,                           
        left=1mm, right=1mm, top=1mm, bottom=1mm, 
        boxsep=4pt,                        
        before skip=10pt, after skip=10pt, 
        title={\faDatabase~~~\textbf{Prompt: Layout Balance Judge}},
        coltitle=white,
        fonttitle=\bfseries\small
    }
}

\tcbset{prompt_aethetic_engagement/.style={
        enhanced,
        breakable,
        colback=prompt,        
        colframe=prompt-frame,            
        fontupper=\normalsize,               
        boxrule=1pt,                       
        arc=4mm,                           
        left=1mm, right=1mm, top=1mm, bottom=1mm, 
        boxsep=4pt,                        
        before skip=10pt, after skip=10pt, 
        title={\faDatabase~~~\textbf{Prompt: Engagement Judge}},
        coltitle=white,
        fonttitle=\bfseries\small
    }
}

\tcbset{prompt_info_clarity/.style={
        enhanced,
        breakable,
        colback=prompt,        
        colframe=prompt-frame,            
        fontupper=\normalsize,               
        boxrule=1pt,                       
        arc=4mm,                           
        left=1mm, right=1mm, top=1mm, bottom=1mm, 
        boxsep=4pt,                        
        before skip=10pt, after skip=10pt, 
        title={\faDatabase~~~\textbf{Prompt: Clarity Judge}},
        coltitle=white,
        fonttitle=\bfseries\small
    }
}

\tcbset{prompt_info_content/.style={
        enhanced,
        breakable,
        colback=prompt,        
        colframe=prompt-frame,            
        fontupper=\normalsize,               
        boxrule=1pt,                       
        arc=4mm,                           
        left=1mm, right=1mm, top=1mm, bottom=1mm, 
        boxsep=4pt,                        
        before skip=10pt, after skip=10pt, 
        title={\faDatabase~~~\textbf{Prompt: Content Completeness Judge}},
        coltitle=white,
        fonttitle=\bfseries\small
    }
}

\tcbset{prompt_info_logic/.style={
        enhanced,
        breakable,
        colback=prompt,        
        colframe=prompt-frame,            
        fontupper=\normalsize,               
        boxrule=1pt,                       
        arc=4mm,                           
        left=1mm, right=1mm, top=1mm, bottom=1mm, 
        boxsep=4pt,                        
        before skip=10pt, after skip=10pt, 
        title={\faDatabase~~~\textbf{Prompt: Logical Flow Judge}},
        coltitle=white,
        fonttitle=\bfseries\small
    }
}

\tcbset{prompt_verbatim_question/.style={
        enhanced,
        breakable,
        colback=prompt,        
        colframe=prompt-frame,            
        fontupper=\normalsize,               
        boxrule=1pt,                       
        arc=4mm,                           
        left=1mm, right=1mm, top=1mm, bottom=1mm, 
        boxsep=4pt,                        
        before skip=10pt, after skip=10pt, 
        title={\faDatabase~~~\textbf{Prompt: Generate Verbatim QA}},
        coltitle=white,
        fonttitle=\bfseries\small
    }
}

\tcbset{prompt_interpretive_question/.style={
        enhanced,
        breakable,
        colback=prompt,        
        colframe=prompt-frame,            
        fontupper=\normalsize,               
        boxrule=1pt,                       
        arc=4mm,                           
        left=1mm, right=1mm, top=1mm, bottom=1mm, 
        boxsep=4pt,                        
        before skip=10pt, after skip=10pt, 
        title={\faDatabase~~~\textbf{Prompt: Generate Interpretive QA}},
        coltitle=white,
        fonttitle=\bfseries\small
    }
}

\tcbset{prompt_answer_agent/.style={
        enhanced,
        breakable,
        colback=prompt,        
        colframe=prompt-frame,            
        fontupper=\normalsize,               
        boxrule=1pt,                       
        arc=4mm,                           
        left=1mm, right=1mm, top=1mm, bottom=1mm, 
        boxsep=4pt,                        
        before skip=10pt, after skip=10pt, 
        title={\faDatabase~~~\textbf{Prompt: Answer Questions}},
        coltitle=white,
        fonttitle=\bfseries\small
    }
}

\tcbset{prompt_4o_image/.style={
        enhanced,
        breakable,
        colback=prompt,        
        colframe=prompt-frame,            
        fontupper=\normalsize,               
        boxrule=1pt,                       
        arc=4mm,                           
        left=1mm, right=1mm, top=1mm, bottom=1mm, 
        boxsep=4pt,                        
        before skip=10pt, after skip=10pt, 
        title={\faDatabase~~~\textbf{Prompt: \texttt{4o-Image}}},
        coltitle=white,
        fonttitle=\bfseries\small
    }
}

\tcbset{prompt_owl_4o/.style={
        enhanced,
        breakable,
        colback=prompt,        
        colframe=prompt-frame,            
        fontupper=\normalsize,               
        boxrule=1pt,                       
        arc=4mm,                           
        left=1mm, right=1mm, top=1mm, bottom=1mm, 
        boxsep=4pt,                        
        before skip=10pt, after skip=10pt, 
        title={\faDatabase~~~\textbf{Prompt: \texttt{OWL-4o}}},
        coltitle=white,
        fonttitle=\bfseries\small
    }
}

\tcbset{prompt_4o_html/.style={
        enhanced,
        breakable,
        colback=prompt,        
        colframe=prompt-frame,            
        fontupper=\normalsize,               
        boxrule=1pt,                       
        arc=4mm,                           
        left=1mm, right=1mm, top=1mm, bottom=1mm, 
        boxsep=4pt,                        
        before skip=10pt, after skip=10pt, 
        title={\faDatabase~~~\textbf{Prompt: \texttt{4o-HTML}}},
        coltitle=white,
        fonttitle=\bfseries\small
    }
}

\tcbset{prompt_parser_summarizer/.style={
        enhanced,
        breakable,
        colback=prompt,        
        colframe=prompt-frame,            
        fontupper=\normalsize,               
        boxrule=1pt,                       
        arc=4mm,                           
        left=1mm, right=1mm, top=1mm, bottom=1mm, 
        boxsep=4pt,                        
        before skip=10pt, after skip=10pt, 
        title={\faDatabase~~~\textbf{Prompt: Paper Summarizer}},
        coltitle=white,
        fonttitle=\bfseries\small
    }
}

\tcbset{prompt_parser_filter/.style={
        enhanced,
        breakable,
        colback=prompt,        
        colframe=prompt-frame,            
        fontupper=\normalsize,               
        boxrule=1pt,                       
        arc=4mm,                           
        left=1mm, right=1mm, top=1mm, bottom=1mm, 
        boxsep=4pt,                        
        before skip=10pt, after skip=10pt, 
        title={\faDatabase~~~\textbf{Prompt: Figure Filter}},
        coltitle=white,
        fonttitle=\bfseries\small
    }
}

\tcbset{prompt_planner_matching/.style={
        enhanced,
        breakable,
        colback=prompt,        
        colframe=prompt-frame,            
        fontupper=\normalsize,               
        boxrule=1pt,                       
        arc=4mm,                           
        left=1mm, right=1mm, top=1mm, bottom=1mm, 
        boxsep=4pt,                        
        before skip=10pt, after skip=10pt, 
        title={\faDatabase~~~\textbf{Prompt: Asset Matching}},
        coltitle=white,
        fonttitle=\bfseries\small
    }
}

\tcbset{prompt_planner_painter/.style={
        enhanced,
        breakable,
        colback=prompt,        
        colframe=prompt-frame,            
        fontupper=\normalsize,               
        boxrule=1pt,                       
        arc=4mm,                           
        left=1mm, right=1mm, top=1mm, bottom=1mm, 
        boxsep=4pt,                        
        before skip=10pt, after skip=10pt, 
        title={\faDatabase~~~\textbf{Prompt: Painter}},
        coltitle=white,
        fonttitle=\bfseries\small
    }
}

\tcbset{prompt_planner_commenter/.style={
        enhanced,
        breakable,
        colback=prompt,        
        colframe=prompt-frame,            
        fontupper=\normalsize,               
        boxrule=1pt,                       
        arc=4mm,                           
        left=1mm, right=1mm, top=1mm, bottom=1mm, 
        boxsep=4pt,                        
        before skip=10pt, after skip=10pt, 
        title={\faDatabase~~~\textbf{Prompt: Commenter}},
        coltitle=white,
        fonttitle=\bfseries\small
    }
}

\tcbset{
    vlm_aesthetics_judge/.style={
        enhanced,
        breakable,
        colback=prompt,        
        colframe=prompt-frame,            
        fontupper=\normalsize,               
        boxrule=1pt,                       
        arc=4mm,                           
        left=1mm, right=1mm, top=1mm, bottom=1mm, 
        boxsep=4pt,                        
        before skip=10pt, after skip=10pt, 
        title={\faDatabase~~~\textbf{Prompt: Aesthetics Quality Judge}},
        coltitle=white,
        fonttitle=\bfseries\small
    }
}
\tcbset{
    vlm_element_judge/.style={
        enhanced,
        breakable,
        colback=prompt,        
        colframe=prompt-frame,            
        fontupper=\normalsize,               
        boxrule=1pt,                       
        arc=4mm,                           
        left=1mm, right=1mm, top=1mm, bottom=1mm, 
        boxsep=4pt,                        
        before skip=10pt, after skip=10pt, 
        title={\faDatabase~~~\textbf{Prompt: Element Quality Judge}},
        coltitle=white,
        fonttitle=\bfseries\small
    }
}

\tcbset{
    vlm_layout_judge/.style={
        enhanced,
        breakable,
        colback=prompt,        
        colframe=prompt-frame,            
        fontupper=\normalsize,               
        boxrule=1pt,                       
        arc=4mm,                           
        left=1mm, right=1mm, top=1mm, bottom=1mm, 
        boxsep=4pt,                        
        before skip=10pt, after skip=10pt, 
        title={\faDatabase~~~\textbf{Prompt: Layout Quality Judge}},
        coltitle=white,
        fonttitle=\bfseries\small
    }
}

\tcbset{
    rpt_prompt/.style={
        enhanced,
        breakable,
        colback=prompt,
        colframe=prompt-frame,
        fontupper=\normalsize,
        boxrule=1pt,
        arc=4mm,
        left=1mm, right=1mm, top=1mm, bottom=1mm,
        boxsep=4pt,
        before skip=10pt, after skip=10pt,
        title={\faDatabase~~~\textbf{Prompt: RPT (Token-Level Reconstruction)}},
        coltitle=white,
        fonttitle=\bfseries\small
    }
}

\tcbset{
    rlpt_prompt/.style={
        enhanced,
        breakable,
        colback=prompt,
        colframe=prompt-frame,
        fontupper=\normalsize,
        boxrule=1pt,
        arc=4mm,
        left=1mm, right=1mm, top=1mm, bottom=1mm,
        boxsep=4pt,
        before skip=10pt, after skip=10pt,
        title={\faDatabase~~~\textbf{Prompt: RLPT (Sentence-Level Reconstruction)}},
        coltitle=white,
        fonttitle=\bfseries\small
    }
}

\tcbset{
    rlvr_prompt/.style={
        enhanced,
        breakable,
        colback=prompt,
        colframe=prompt-frame,
        fontupper=\normalsize,
        boxrule=1pt,
        arc=4mm,
        left=1mm, right=1mm, top=1mm, bottom=1mm,
        boxsep=4pt,
        before skip=10pt, after skip=10pt,
        title={\faDatabase~~~\textbf{Prompt: RLVR}},
        coltitle=white,
        fonttitle=\bfseries\small
    }
}
\tcbset{prompt_analyze/.style={
    enhanced,
    colback=prompt!25, colframe=prompt-frame, fontupper=\small,
    boxrule=1pt, arc=4mm, left=1mm, right=1mm, top=1mm, bottom=1mm, boxsep=4pt,
    before skip=10pt, after skip=10pt,
    title={\faDatabase~~~\textbf{Prompt: Analyze stage}},
    coltitle=white, fonttitle=\bfseries\small
}}
\tcbset{prompt_plan/.style={
    enhanced,
    colback=prompt!25, colframe=prompt-frame, fontupper=\small,
    boxrule=1pt, arc=4mm, left=1mm, right=1mm, top=1mm, bottom=1mm, boxsep=4pt,
    before skip=10pt, after skip=10pt,
    title={\faDatabase~~~\textbf{Prompt: Plan stage}},
    coltitle=white, fonttitle=\bfseries\small
}}
\tcbset{prompt_produce/.style={
    enhanced,
    colback=prompt!25, colframe=prompt-frame, fontupper=\small,
    boxrule=1pt, arc=4mm, left=1mm, right=1mm, top=1mm, bottom=1mm, boxsep=4pt,
    before skip=10pt, after skip=10pt,
    title={\faDatabase~~~\textbf{Prompt: Produce stage}},
    coltitle=white, fonttitle=\bfseries\small
}}
\tcbset{prompt_controller/.style={
    enhanced,
    colback=prompt!25, colframe=prompt-frame, fontupper=\small,
    boxrule=1pt, arc=4mm, left=1mm, right=1mm, top=1mm, bottom=1mm, boxsep=4pt,
    before skip=10pt, after skip=10pt,
    title={\faDatabase~~~\textbf{Prompt: Observation controller (lever selection)}},
    coltitle=white, fonttitle=\bfseries\small
}}
\tcbset{prompt_harness/.style={
    enhanced,
    colback=prompt!25, colframe=prompt-frame, fontupper=\small,
    boxrule=1pt, arc=4mm, left=1mm, right=1mm, top=1mm, bottom=1mm, boxsep=4pt,
    before skip=10pt, after skip=10pt,
    title={\faDatabase~~~\textbf{Prompt: Harness-evolution proposer}},
    coltitle=white, fonttitle=\bfseries\small
}}

\newtcblisting{tcbverbatim}{
    listing only,
    listing options={basicstyle=\ttfamily},
    colback=gray!10,
    colframe=gray!50,
    breakable
}

\newtcolorbox{examplebox}{
  enhanced,
  breakable,
  colback=gray!3,
  colframe=black,
  boxrule=0.8pt,
  arc=2pt,
  left=6pt,
  right=6pt,
  top=6pt,
  bottom=6pt
}

\title{Sage-Evolve-Preview: \\ Adaptive Observation for Post-Training Self-Improvement}

\author{
    \textbf{Yinhao Tang}$^{1,2}$\textsuperscript{*} \quad
    \textbf{Youqing Fang}$^{1,2}$\textsuperscript{*} \quad
    \textbf{Yanan Sun}$^{2}$\textsuperscript{\dag} \quad   \\
    \textbf{Xintian Shen}$^{2}$\quad
    \textbf{Hanyu Sun}$^{2}$\quad 
    \textbf{Wenran Liu}$^{2}$\quad 
    \textbf{Yue Wu}$^{2}$ \quad \\
    \textbf{Weiming Zhang}$^{1}$\quad 
    \textbf{Bin Liu}$^{1}$ \quad
    \textbf{Wenwei Zhang}$^{2}$ \quad \\
    \textsuperscript{1}University of Science and Technology of China \textsuperscript{2}Shanghai AI Laboratory \\
    \normalsize
    \normalfont
    \texttt{\{tangyinhao,fangyq\}@mail.ustc.edu.cn}, 
    \texttt{\{sunyanan\}@pjlab.org.cn}
}

\begin{document}
\pagestyle{plain}
\maketitle

\begingroup
\renewcommand{\thefootnote}{}
\footnotetext{* Equal contribution.\, $\dag$ Corresponding author.}
\endgroup

\begin{abstract}
Post-training recursive self-improvement (RSI) iteratively improves a target capability, operationalized by a fixed anchor benchmark, through repeated intervention, training, and evaluation.
Yet most loops use the anchor's aggregate score as their sole observation, so early gains give way to stalled progress and repetitive exploration.
When this anchor-only feedback ceases to reveal actionable gaps, a plateau may reflect an exhausted observation rather than an exhausted capability.
We introduce \emph{adaptive observation} (AdaObs), an observation-side mechanism of harness evolution that keeps the anchor objective fixed while adapting the active observation.
At a plateau, a controller refines diagnostic granularity, escalates difficulty, and adds related tasks; validated increments are fused back into training, while every checkpoint remains evaluated against the anchor.
Across eight independently optimized anchors in long-context reading comprehension and instruction following, AdaObs raises the mean score from $66.4$ to $73.7$ under a matched data budget, outperforming FixedObs, the anchor-only baseline.
The evolved harness further improves the seven-task average on the independent PostTrainBench across two agent backends.
These results identify observation adaptation as a missing mechanism of harness evolution for sustaining post-training RSI.

\end{abstract}

\begin{figure}[t]
    \centering
    \includegraphics[width=\columnwidth]{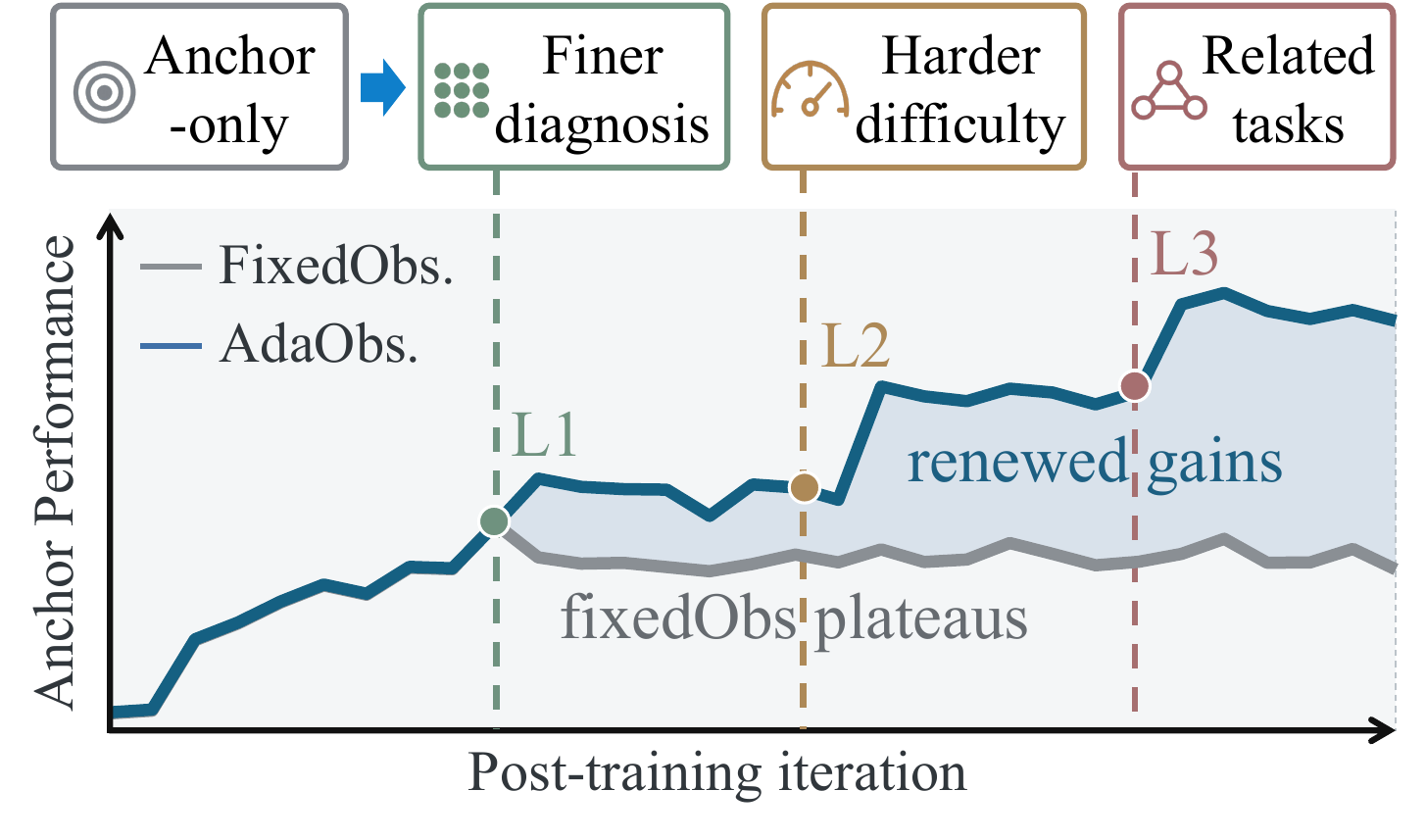}
    \vspace{-0.45em}
    \caption{
    \textbf{Adaptive observation sustains recursive self-improvement.}
    FixedObs (gray), which relies on anchor-only feedback, plateaus; AdaObs (blue) adapts the active observation by refining diagnosis (L1), escalating difficulty (L2), and adding related tasks (L3), before fusing validated gains back.
    }
    \label{fig:teaser}
    \vspace{-0.35em}
\end{figure}

\section{Introduction}
\label{sec:intro}

Post-training increasingly takes the form of a feedback-driven search over data and training decisions rather than a fixed recipe~\citep{jaech2024openai,guo2025deepseek,bai2026intern}. In this framing, an agent drives the search: it inspects model failures, proposes a targeted intervention, trains a new checkpoint, and reads benchmark outcomes to decide what to try next. Because training and evaluation are repeatable, the agent can run this cycle many times, each result sharpening the next intervention. We study this closed-loop process as recursive self-improvement (RSI): each run targets a designated model capability, a fixed benchmark anchors the objective, and experimental feedback connects successive interventions. By accumulating validated gains across rounds, such a loop can keep finding productive steps even after the recipe has exhausted its most obvious opportunities.

In practice, however, autonomous post-training loops often lose momentum. Recent studies report that early gains give way to stalled progress and repetitive exploration: agents keep launching near-identical training variants on a plateaued benchmark instead of rethinking their approach, so that ``execution is solved, rethinking is not''~\citep{lim2026missing, rank2026posttrainbench}. In the conventional setting, the same anchor serves two roles: it defines the fixed optimization objective and provides the loop's sole observation. We refer to this anchor-only setting as \textbf{FixedObs}. Its aggregate score can indicate whether an intervention helped, but often not which capability gap remains or what intervention should come next. Existing systems mainly evolve what the loop acts on or how it acts, including training data~\citep{du2026datamaster, wang2026curateevo}, synthesis code~\citep{zhang2026self}, and agent-side harness components~\citep{lee2026meta}, while keeping the observation used for experiment selection fixed. We identify this as a missing dimension of harness evolution: when anchor-only feedback ceases to reveal actionable gaps, a plateau may reflect an exhausted observation rather than an exhausted capability.

We therefore decouple the fixed anchor objective from the active observation used to guide optimization. We introduce \emph{adaptive observation} (\textbf{AdaObs}), an observation-side mechanism of harness evolution. Every checkpoint remains evaluated against the anchor, but when progress plateaus, a controller adapts the diagnostic evidence that guides the next intervention through an ordered ladder: (L1)~refining \emph{diagnostic granularity} to expose failure modes hidden by the aggregate score, (L2)~escalating to harder \emph{difficulty levels} that restore headroom, and (L3)~adding \emph{related tasks} that reveal transferable gaps. The controller consults experiment memory and skips concrete updates that have already failed. Validated increments discovered under these cumulative observations are merged into the retained training set within each run, redirecting optimization toward the anchor without moving the objective (Figure~\ref{fig:teaser}).

We realize this policy as a fully autonomous loop on production training infrastructure, running repeated full-parameter SFT. Across long-context reading comprehension and instruction following, we optimize eight anchor benchmarks independently. With the same starting data and training recipe and a comparable data budget, AdaObs outperforms FixedObs on all eight, raising their mean score from $66.4$ to $73.7$. The evolved harness also transfers unchanged to PostTrainBench, improving its seven-task average by up to $6.9$ points.

Our analyses expose the mechanism behind these gains. Diagnostic refinement (L1) makes heterogeneous failures visible: decomposing aggregate feedback into behavioral indicators reveals a supervision-format mismatch that drives over-length, truncated answers. Difficulty escalation (L2) moves the active observation to harder subsets, whose increments transfer asymmetrically back to easier subsets and the anchor. Lateral expansion (L3) introduces complementary tasks whose supervision continues to improve the anchor after anchor-only optimization has saturated. Together, these results show why no single observation remains uniformly informative: progress resumes when the harness shifts to an observation that reveals a new, transferable gap. On the agent side, persistent harness updates also reduce recurrent failures such as capability misdiagnosis, data contamination, and redundant exploration across both backends.
To summarize, our key contributions are as follows:
\begin{itemize}[leftmargin=1.4em,itemsep=2pt,topsep=2pt]
    \item We identify \emph{observation adaptation}, evolving the diagnostic evidence used to select interventions while preserving a fixed anchor objective, as an overlooked observation-side mechanism of harness evolution.
    \item We instantiate this principle as a three-level ladder of diagnostic refinement (L1), difficulty escalation (L2), and cumulative expansion to related tasks (L3), coordinated by a plateau-triggered controller that accumulates validated increments and avoids failed interventions through experiment memory.
    \item Across eight separately optimized anchors, AdaObs sustains gains beyond anchor-only feedback and the evolved harness transfers unchanged to PostTrainBench across tasks and agent backends, with analyses showing that each lever resumes progress by exposing a new, transferable capability gap.
\end{itemize}

\section{Related Work}
\label{sec:related}

\paragraph{Autonomous and agentic post-training.}
Recent work asks whether LLM agents can automate post-training. PostTrainBench~\citep{rank2026posttrainbench} tasks agents with budgeted research, data synthesis, and training-code iteration, documenting failures such as contamination and reward hacking. Data-centric systems optimize this loop: DataMaster~\citep{du2026datamaster} tree-searches over curation strategies, CurateEvo~\citep{wang2026curateevo} rewrites policies from failed trajectories, LaMDAgent~\citep{yano2025lamdagent} searches over pipelines, and LoopTool~\citep{zhang2026looptool} closes a probe-synthesize-retrain loop for tool use. Across 1{,}338 trajectories, agents master execution but rarely re-assess strategy at plateaus~\citep{lim2026missing}: ``execution is solved, rethinking is not.'' These systems evolve data, curation code, or pipelines while holding evaluation fixed. We instead adapt the observation to escape these plateaus.

\paragraph{Self-improving and self-evolving agents.}
Beyond data, many systems recursively revise their own artifacts: self-improving harnesses~\citep{zhang2026self, lee2026meta}, self-modifying agent code~\citep{zhang2026darwin}, and evolving skills, memory, or context~\citep{wang2023voyager, ouyang2509reasoningbank}; SEAL~\citep{zweiger2026self} lets a model generate its own finetuning data and self-edits. A common choice is to keep the evaluator outside the evolving loop to prevent reward hacking~\citep{lin2026agentic}. Our work is complementary: we fix the objective and its verification discipline but let the loop adapt what it observes, at what granularity, at what difficulty, and over which tasks.

\paragraph{Evolving evaluation, curricula, and difficulty.}
Two adjacent halves of the literature touch our idea. On the measurement side, benchmarks are made dynamic to resist saturation: AutoBencher~\citep{li2407autobencher} searches for harder benchmarks, ArenaBencher~\citep{liu2025arenabencher} evolves items via multi-model competition, and DyVal~\citep{zhu2309dyval} generates controllable-difficulty samples, while nearly half of surveyed benchmarks saturate as models improve~\citep{akhtar2602ai}. On the training side, self-play and curriculum systems adapt what to train on: Absolute Zero~\citep{zhao2026absolute}, R-Zero~\citep{huang2026r}, 
Tome-Math~\citep{fang2026scaling},
SPICE~\citep{liu2510spice}, and SCOPE~\citep{kwan2026scope} generate tasks at the learner's frontier, and rubric- or reward-evolving methods co-evolve the reward with the policy~\citep{iacob2026red, shao2511dr}. The former target leaderboard comparability; the latter evolve the training distribution or reward directly, whereas we adapt the observation that selects which increments to train on while keeping the anchor objective fixed. We unite these halves inside a single RSI loop. Because fixed feedback loses information as models improve~\citep{devilling2025mirror, chen2026rethinking}, a plateau triggers adaptation of the observation, across granularity, difficulty, and task set, to expose hidden capability gaps and re-open gains on the target capability.

\section{Method}
\label{sec:method}

We cast post-training as a recursive self-improvement (RSI) loop that iteratively trains and re-evaluates a model, linking runs through an experiment memory (\S\ref{sec:method:loop}). The loop stalls when its feedback goes stale: a plateau may reflect an exhausted capability, but it may also mean that the loop is viewing that capability through an uninformative lens. We address the latter case by evolving two parts of the loop on different timescales---the active observation changes when progress plateaus (\S\ref{sec:method:adapt}), while the agent harness changes when recurring failures surface (\S\ref{sec:method:harness}).

\begin{figure*}[t]
    \centering
    \includegraphics[width=\textwidth]{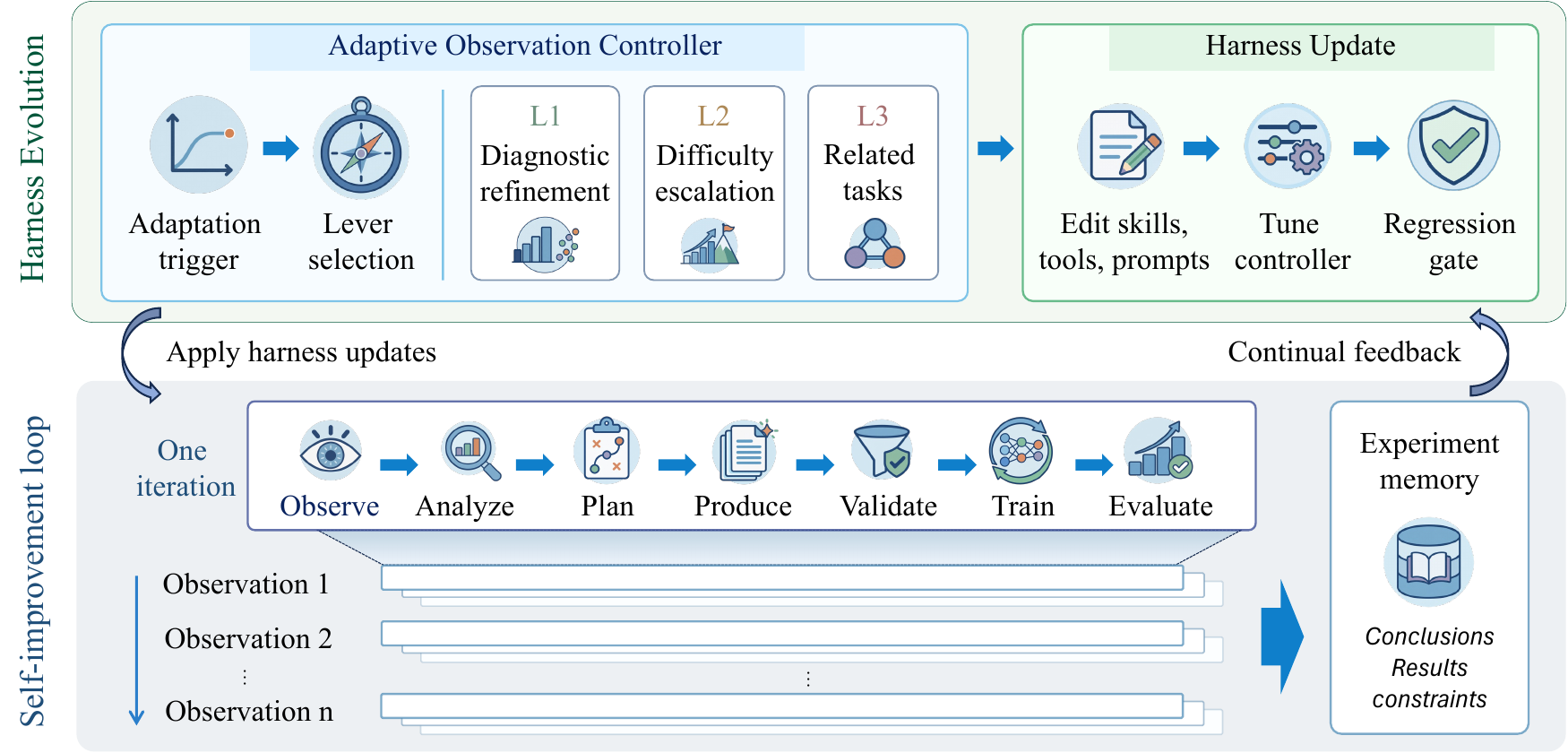}
    \caption{
    \textbf{Overview of the self-evolving post-training harness.}
    Each iteration runs under an active observation and records its outcome
    in experiment memory. At a plateau, the controller climbs an ordered
    observation ladder from finer diagnosis to harder instances and then to
    related tasks. On a slower timescale, recurring agent failures trigger
    bounded harness edits that must pass replay-based regression tests.
    }
    \label{fig:pipeline}
\end{figure*}

\subsection{The self-improvement loop}
\label{sec:method:loop}

We optimize a base model $M_0$ toward target capabilities $\mathcal{C}$. Each iteration executes seven stages: \textsc{Observe} collects feedback from the latest checkpoint under the active observation; \textsc{Analyze} identifies a capability gap; \textsc{Plan} turns it into a minimal, falsifiable data intervention; \textsc{Produce} generates the corresponding SFT increment by synthesizing new data or downloading a public dataset; \textsc{Validate} applies quality, provenance, and contamination gates; \textsc{Train} runs full-parameter SFT on the base data plus all retained increments; and \textsc{Evaluate} scores the resulting checkpoint. Successive iterations are connected by the evaluated checkpoint and an experiment memory $\mathcal{M}$, which stores confirmed conclusions and negative results so the loop does not repeatedly propose interventions that have already failed. This closes the loop: each checkpoint seeds the next iteration, and $\mathcal{M}$ carries forward what earlier rounds learned.

An observation is $O_r=(\mathcal{B}_r,\delta_r,g_r)$, where $\mathcal{B}_r$ is the benchmark set, $\delta_r$ selects its active tiers or subsets, and $g_r$ controls the diagnostic resolution. We distinguish a fixed anchor observation $O^{\mathrm{anc}}$, the original target, from the active observation $O_r$ guiding phase $r$; every checkpoint is measured on both, the active score testing the current hypothesis and the anchor score guarding the unchanged objective. An increment is retained if it improves either score within a small anchor-regression tolerance, accumulating in $\Delta^\star$.

Within a phase, $O_r$ is fixed and we normalize improvement by the amount of added data:
\begin{equation}
\label{eq:mr}
\mathrm{MR}_t=
\frac{\textsc{Score}(M_t;O_r)-\textsc{Score}(M_{t-1};O_r)}
{\Delta d_t}.
\end{equation}
We exponentially smooth this marginal return and declare a plateau when every value in a trailing window of length $w$ is below $\epsilon$:
\begin{equation}
\label{eq:plateau}
\begin{aligned}
\widehat{\mathrm{MR}}_t
&=\alpha\mathrm{MR}_t+(1-\alpha)\widehat{\mathrm{MR}}_{t-1},\\
\operatorname{Plateau}(t)
&\Longleftrightarrow
\max_{t-w<i\le t}\widehat{\mathrm{MR}}_i<\epsilon.
\end{aligned}
\end{equation}
Changing the active observation starts a new phase and resets this history, so scores from different observations are never compared directly.

\begin{table*}[!t]
\centering
\caption{
\textbf{Main results.} Columns are benchmarks, each optimized separately as its own anchor. For FixedObs and AdaObs each cell is the best checkpoint of an independent per-benchmark loop, so a row collects per-benchmark results rather than one model scored across all columns; top-block reference models are single checkpoints run on every benchmark. \textbf{Avg.}\ is the mean over the eight benchmarks. \besti{Best} and \secondi{second best} per column are highlighted.
}
\label{tab:main}
\small
\setlength{\tabcolsep}{3pt}
\resizebox{\textwidth}{!}{
\begin{tabular}{lccccccccc}
\toprule
\multirow{2}{*}{\textbf{Model}} & \multicolumn{4}{c}{\textbf{Reading Comprehension}} & \multicolumn{4}{c}{\textbf{Instruction Following}} & \multirow{2}{*}{\textbf{Avg.}} \\
\cmidrule(lr){2-5} \cmidrule(lr){6-9}
 & \makecell{LongBench\\v2} & \makecell{MRCR v2\\upto256k} & AA-LCR & SummHay & \makecell{Advanced\\IF} & IFBench & LIFBench & \makecell{LongGen\\Bench} & \\
\midrule
\multicolumn{10}{l}{\textit{Reference models}} \\
\quad Qwen3.5-35B-A3B-base & 52.09 & 24.17 & 44.00 & 8.11 & 58.94 & 46.06 & 79.00 & 51.00 & 45.42 \\
\quad Qwen3.5-35B-A3B      & \secondi{58.05} & 63.26 & \secondi{67.00} & 26.07 & 67.47 & \secondi{66.66} & \secondi{92.55} & 61.00 & 62.76 \\
\quad Qwen3.6-35B-A3B      & 55.07 & 83.94 & 66.00 & 27.41 & \besti{73.21} & 59.20 & 92.00 & 64.52 & 65.17 \\
\quad base + 450k & 52.24 & 29.60 & 56.67 & 16.19 & 61.27 & 58.79 & 82.42 & 59.50 & 52.09 \\
\midrule
\multicolumn{10}{l}{\textit{This work}} \\
\quad FixedObs       & 55.00 & \secondi{87.20} & 66.00 & \secondi{38.69} & 63.12 & 60.51 & 89.76 & \secondi{71.26} & \secondi{66.44} \\
\rowcolor{oursrow}
\quad \textbf{AdaObs (ours)}   & \besti{58.12} & \besti{95.57} & \besti{72.00} & \besti{50.07} & \secondi{68.86} & \besti{66.88} & \besti{93.61} & \besti{84.19} & \besti{73.66} \\
\bottomrule
\end{tabular}
}
\end{table*}

\subsection{Adaptive observation controller}
\label{sec:method:adapt}

At a plateau, the controller advances along the fixed ladder $\textsc{L1}\prec\textsc{L2}\prec\textsc{L3}$, taking the first feasible lever and skipping concrete updates already ruled out by $\mathcal{M}$. The ladder fixes only the order in which levers are tried; which concrete update to apply within a lever, that is, which indicators to decompose the score into, which harder tier to escalate to, or which related benchmark to add, is the controller's own choice at run time. If $T_\ell$ denotes the update associated with lever $\ell$, then $O_{r+1}=T_\ell(O_r)$, where
\begin{equation}
T_\ell(O_r)=
\begin{cases}
(\mathcal{B}_r,\delta_r,\operatorname{Refine}(g_r)), & \ell=\textsc{L1},\\
(\mathcal{B}_r,\delta_r\cup\{\delta^\star\},g_r), & \ell=\textsc{L2},\\
(\mathcal{B}_r\cup\{B^\star\},\delta_r,g_r), & \ell=\textsc{L3}.
\end{cases}
\label{eq:lever-transitions}
\end{equation}
Thus each step changes one component of the observation while inheriting the other two.

\paragraph{L1: diagnostic refinement.}
The controller first keeps the task and difficulty fixed and refines only $g_r$. It decomposes an aggregate score into item-level failure groups, protocol splits, or behavioral indicators such as over-length and think-trigger ratios, then reruns \textsc{Analyze} over the same evaluation outputs. This is the cheapest step because it can reveal an actionable defect without collecting new evaluation data.

\paragraph{L2: difficulty escalation.}
If finer diagnosis yields no new gap, the controller keeps the task identity fixed and adds a harder difficulty tier $\delta^\star$ of its own choosing to the active observation, using a held-out reference model as a guide so the chosen tier stays within an informative pass-rate band: hard enough to provide headroom, but not so hard that it ceases to provide useful feedback. The previously observed tiers are retained alongside it, so the observation accumulates difficulty rather than replacing it. The selection rule is given in Appendix~\ref{sec:appendix:observation}.

\paragraph{L3: lateral expansion.}
If the within-benchmark options are exhausted, the controller selects an untried benchmark $B^\star$ in the same capability direction but with a different task form and adds it to the active set, $\mathcal{B}_{r+1}=\mathcal{B}_r\cup\{B^\star\}$. The new task supplies the next diagnostic focus, while previously added tasks stay in the cumulative observation and the target remains fixed in $O^{\mathrm{anc}}$. This lateral view can expose transferable sub-skills the original task did not isolate, without discarding earlier evidence.

\subsection{Harness update}
\label{sec:method:harness}

Observation adaptation changes what the loop looks at; harness updates change how it responds. On a slower clock, the system mines $\mathcal{M}$ for recurring agent-side failures, such as misdiagnosis, data contamination, redundant exploration, or skipped validation. A supported failure triggers several bounded candidates, each modifying one editable surface of the agent harness: a stage prompt for \textsc{Analyze}, \textsc{Plan}, or \textsc{Produce}, a skill, a tool interface, a memory rule, or an observation-controller rule.

Each candidate is replayed on two sets of logged iterations: a repair set containing the targeted failure and a regression buffer containing previously successful decisions. The best candidate is committed only if it repairs the target cases, preserves the regression buffer, and passes interface, provenance, and prompt-budget checks; otherwise the active harness is unchanged. The evaluator, anchor observation, and validation gates remain immutable, preventing the harness from improving by changing its own test. Its formal acceptance rule and the full controller appear in Appendix~\ref{sec:appendix:harness}.

A deterministic execution layer submits distributed training and evaluation jobs, validates checkpoints, and persists state for failure recovery. It makes no scientific decisions, but enables the loop to run unattended on a production cluster; implementation details are deferred to Appendix~\ref{sec:appendix:cluster}.

\begin{figure*}[t]
\centering
\includegraphics[width=\textwidth]{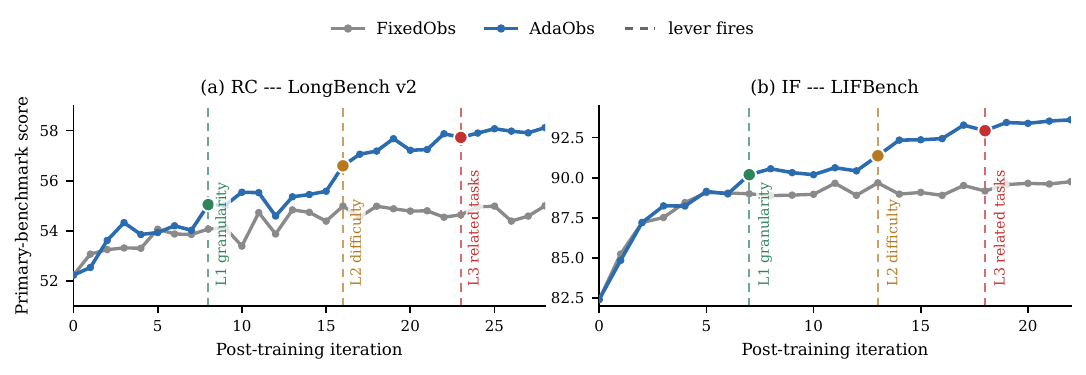}
\caption{
\textbf{Adaptive observation sustains gains past the fixed-observation plateau.}
Per-iteration score on two example anchor benchmarks, (a)~LongBench~v2 and (b)~LIFBench. The grey curve holds a fixed observation and plateaus; AdaObs (blue) re-opens marginal returns in three lever stages marked by dashed verticals: L1 diagnostic granularity, L2 difficulty, and L3 related tasks. Both curves end at the Table~\ref{tab:main} scores.
}
\label{fig:main-exp}
\end{figure*}

\begin{table*}[t]
\centering
\caption{
\textbf{PostTrainBench transfer: our evolved harness vs.\ the default harness under two agent backends.} Seven sub-benchmarks span math, code, science, tool use, medicine, and creative writing. Each cell is the mean over four post-trained student models, reported as avg@4 per model (details in \S\ref{sec:exp:ptb}).
}
\label{tab:ptb}
\small
\setlength{\tabcolsep}{5pt}
\resizebox{\textwidth}{!}{
\begin{tabular}{llcccccccc}
\toprule
\textbf{Harness} & \textbf{Agent backend} & \makecell{AIME\\2025} & GSM8K & GPQA & HumanEval & BFCL & \makecell{ArenaHard\\-Writing} & \makecell{HealthBench\\-Easy} & \textbf{Avg.} \\
\midrule
Base model      & --               &  5.1 & 45.0 & 22.6 & 31.5 &  1.7 &  7.3 & 19.1 & 18.9 \\
\midrule
Default harness & GLM-5.2          &  5.2 & 61.8 & 25.6 & 46.5 & 46.1 &  8.4 & 18.2 & 30.3 \\
\rowcolor{oursrow}
AdaObs (ours)   & GLM-5.2          & \textbf{6.7} & \textbf{69.3} & \textbf{28.8} & \textbf{54.6} & \textbf{70.7} & \textbf{10.7} & \textbf{19.4} & \textbf{37.2} \\
\midrule
Default harness & Claude-Opus-4.8  &  7.9 & 67.8 & 30.2 & 54.6 & 55.1 & 11.7 & 22.6 & 35.7 \\
\rowcolor{oursrow}
AdaObs (ours)   & Claude-Opus-4.8  & \textbf{9.7} & \textbf{73.8} & \textbf{33.2} & \textbf{61.6} & \textbf{78.3} & \textbf{15.1} & \textbf{24.4} & \textbf{42.3} \\
\bottomrule
\end{tabular}
}
\end{table*}

\section{Experiments}
\label{sec:experiments}

\subsection{Setup}
\label{sec:exp:setup}

\paragraph{Base model.}
Our main post-training study starts from Qwen3.5-35B-A3B-base~\citep{qwen35} and uses full-parameter SFT throughout. We optimize each benchmark as its own anchor, running a separate loop that holds that benchmark fixed as the target. Within each loop, the fixed-observation route (\textbf{FixedObs}) and our adaptive-observation route (\textbf{AdaObs}) start from the same data and training recipe. As references we additionally report the base model, the base model fine-tuned on the 450k base increments alone, and two off-the-shelf Qwen checkpoints, Qwen3.5-35B-A3B~\citep{qwen35} and Qwen3.6-35B-A3B~\citep{qwen36_35b_a3b}.

\paragraph{Evaluation.}
We evaluate on two benchmark categories. Reading comprehension is measured by LongBench~v2~\citep{bai2025longbench}, MRCR~v2 upto256k~\citep{deepmind2026mrcr}, AA-LCR~\citep{aa2025lcr}, and SummHay~\citep{laban2024summary}; instruction following by AdvancedIF~\citep{he2026advancedif}, IFBench~\citep{pyatkin2026generalizing}, LIFBench~\citep{wu2025lifbench}, and LongGenBench~\citep{liu2024longgenbench}. On both tracks, FixedObs and AdaObs drive their post-training agent with the Claude-Opus-4.8~\citep{anthropic2026claudeopus48} backend. Beyond these two directions, we test harness-level transfer by freezing the evolved harness and applying it unchanged to PostTrainBench~\citep{rank2026posttrainbench}, whose seven tasks span math, code, science, tool use, medicine, and creative writing. We compare it with the default PostTrainBench harness under two agent backends, GLM-5.2~\citep{glm5team2026glm5vibecodingagentic} and Claude-Opus-4.8~\citep{anthropic2026claudeopus48}.

\subsection{Main results}
\label{sec:exp:main}

\paragraph{Adaptive observation breaks the fixed-observation plateau.}
Table~\ref{tab:main} compares FixedObs against AdaObs from the same starting recipe and a comparable effective-increment budget, each benchmark optimized as its own anchor. The fixed-observation route exhausts its view early and plateaus, whereas AdaObs re-opens progress and reaches a higher asymptote on every benchmark. Averaged within each domain, it lifts reading comprehension from $61.7$ to $68.9$ and instruction following from $71.2$ to $78.4$, raising the mean over the separately optimized benchmarks from $66.4$ to $73.7$ and surpassing both off-the-shelf Qwen checkpoints (Qwen3.5-35B-A3B at $62.8$ and Qwen3.6-35B-A3B at $65.2$). Because both routes start from the same recipe and use a comparable effective-increment budget (per-benchmark budgets in Appendix Table~\ref{tab:increment_budget}), these gains come from changing the observation rather than from adding more of the same data, and they are uniform, widest on the highest-headroom benchmarks LongGenBench ($+12.9$) and SummHay ($+11.4$).


\paragraph{The three levers fire in sequence.}
Figure~\ref{fig:main-exp} traces how the renewed gains arrive, taking LongBench~v2 and LIFBench as two anchors. On each, the FixedObs model warms up and then plateaus, while AdaObs re-opens marginal returns in three stages as the controller fires successive levers: L1 refines the diagnostic granularity, L2 escalates difficulty, and L3 adds a related benchmark to the observation. Each lever produces a fresh increment that, once validated and fused back, lifts the anchor benchmark to a new plateau rather than exhausting it. We dissect the mechanism behind each lever and show the per-subset and per-task curves it induces in \S\ref{sec:analysis}. A single-lever ablation (Appendix Table~\ref{tab:lever_ablation}) confirms that each lever alone recovers only part of the gain, so the full ladder is needed.

\begin{figure*}[t]
    \centering
    \includegraphics[width=\textwidth]{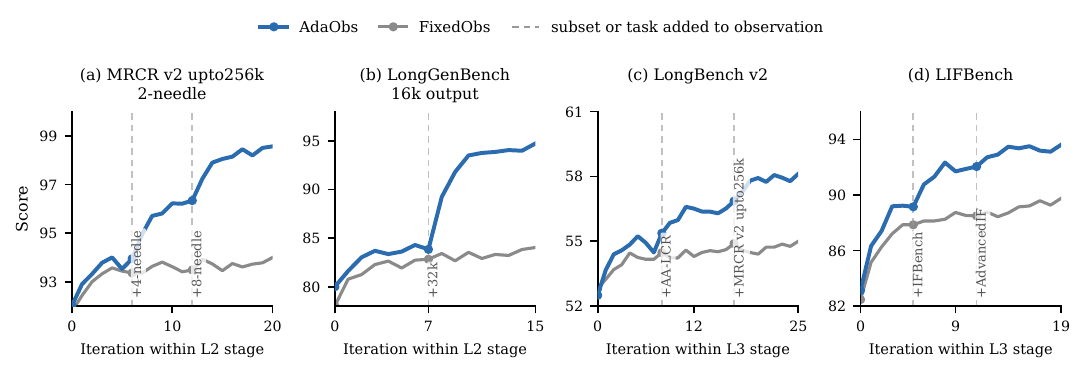}
    \caption{
    \textbf{The difficulty (L2) and related-task (L3) levers each re-open marginal returns.}
    L2 (a,b): on MRCR~v2 upto256k (needle count) and LongGenBench (output length), the earliest-observed subset keeps climbing as harder subsets are added. L3 (c,d): the primary benchmark (LongBench~v2, LIFBench) steps up each time a same-direction benchmark enters its observation. Blue is AdaObs, grey FixedObs; dashed verticals mark each entry. Panels (a,b) plot a single subset, so their axes are subset scores. Grey endpoints and per-stage values are in Appendix~\ref{sec:appendix:difficulty}.
    }
    \label{fig:difficulty}
\end{figure*}

\subsection{Generalization to PostTrainBench}
\label{sec:exp:ptb}

Table~\ref{tab:ptb} reports the PostTrainBench transfer. Crucially, we do not run the initial harness: we take the evolved harness, the one our harness-evolution track (\S\ref{sec:method:harness}) produced after its prompts and controller heuristics had converged on our own post-training runs, and transfer it unchanged. Our evolved harness improves over the default under both backends, lifting the seven-benchmark average from $30.3$ to $37.2$ under GLM-5.2 and from $35.7$ to $42.3$ under Claude-Opus-4.8, which shows that the adaptive-observation machinery, not the backend model, drives the gain.

\section{Understanding Adaptive Observation and Harness Evolution}
\label{sec:analysis}

We next study why the two evolving components of our system are effective. We walk up the observation ladder lever by lever, first showing how refining diagnostic granularity (L1) exposes a format-induced training defect, then how escalating difficulty (L2) and switching to related tasks (L3) each re-open marginal returns. We then ask why this works, by measuring cross-observation transfer. We finally test whether harness evolution reduces recurrent failures of the agent backend.

\subsection{Diagnostic granularity (L1) exposes a format-induced defect}
\label{sec:analysis:cot}

Our first case shows what the cheapest lever buys. While optimizing the two long-context reading benchmarks MRCR~v2 upto256k and AA-LCR, both stalled early with many over-length, truncated answers that the aggregate alone could not attribute to a capability ceiling. Firing L1 decomposes the score into two behavioral indicators: the over-length ratio and the think-trigger ratio, the fraction of items opening a think block. The latter was high but not complete ($0.91$), so the problem was not failing to think but how the thinking unfolded. The agent traced this to the data format: increments carrying no think (CoT) trace, mixed into the think-based 450k base, dilute the think supervision, so on harder items an opened think block rambles without terminating and truncates the answer.

The agent re-rolled every increment to carry a think trace matching the base, changing the supervision format, not its content or volume. Table~\ref{tab:think} shows the over-length ratio dropping from $0.31$ to $0.08$ and the think-trigger ratio reaching $0.99$, which lifts MRCR~v2 upto256k from its $87.20$ plateau to $90.33$ and AA-LCR from $66.00$ to $69.00$. Invisible at the aggregate level, the defect became actionable only once L1 refined the observation into these two indicators.

\begin{table}[t]
\centering
\caption{
\textbf{Think vs.\ no-think increment mixing (RC).} Mixing no-think increments into the think-based 450k base destabilizes reasoning and inflates the over-length ratio; re-rolling them to the think format restores convergence. Think enabled throughout.
}
\label{tab:think}
\small
\setlength{\tabcolsep}{4pt}
\resizebox{\columnwidth}{!}{
\begin{tabular}{lcccc}
\toprule
\makecell[l]{\textbf{Increment}\\(450k think base +)} & \makecell{MRCR v2\\upto256k} & \makecell{AA-\\LCR} & \makecell{Over-len.\\ratio} & \makecell{Think-trig.\\ratio} \\
\midrule
no-think incr. & 87.20 & 66.00 & 0.31 & 0.91 \\
\rowcolor{skyblue!30}
think incr.    & 90.33 & 69.00 & 0.08 & 0.99 \\
\bottomrule
\end{tabular}
}
\end{table}

\subsection{Difficulty (L2) and related tasks (L3) re-open marginal returns}
\label{sec:analysis:difficulty}

The two data-producing levers each re-open marginal returns in their own right (Figure~\ref{fig:difficulty}). The difficulty lever (L2) resolves a benchmark into harder within-benchmark subsets and adds them one at a time: on MRCR~v2 upto256k by needle count and LongGenBench by output length, a subset already under observation keeps improving as later, harder subsets are added, so the earliest-observed subset ends up the strongest of the set. The related-task lever (L3) instead expands the benchmark set: as each same-direction benchmark enters the observation, with AA-LCR then MRCR~v2 upto256k for LongBench~v2 and IFBench then AdvancedIF for LIFBench, the primary benchmark itself steps up again, well past the fixed-observation plateau. In both cases the renewed gains come from changing what the loop observes, not from more of the same data on the original target.

\subsection{Cross-observation transfer and complementarity}
\label{sec:analysis:transfer}

\begin{figure}[t]
    \centering
    \includegraphics[width=\columnwidth]{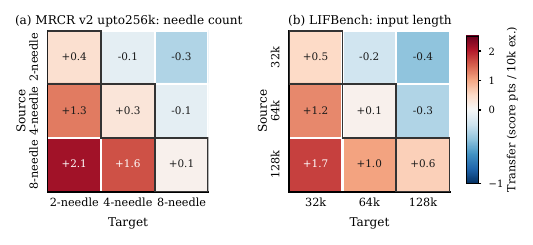}
    \caption{
    \textbf{Cross-observation transfer is asymmetric.}
    Normalized transfer (Eq.~\ref{eq:mr}; score points per $10$k increment examples, see colorbar) from each source subset (row) to each target subset (column) within the same benchmark; the boxed diagonal is self-transfer. See \S\ref{sec:analysis:transfer}.
    }
    \label{fig:transfer}
\end{figure}

Having seen each lever re-open returns, we ask why changing the observation helps at all. We measure how an increment produced for one observation affects the other observations within the same capability direction, summarizing these effects as transfer heatmaps (Figure~\ref{fig:transfer}), with the source subset in rows, the held-out target in columns, and each cell the resulting change per $10$k increment examples (Eq.~\ref{eq:mr}). Two patterns emerge, both across the needle-count (a) and input-length (b) directions. First, transfer is asymmetric: harder subsets lift easier ones (the lower triangle is strongly positive, up to $+2.1$ from 8-needle to 2-needle and $+1.7$ from 128k to 32k), whereas the reverse is weak or negative (down to $-0.4$). Second, an observation saturated on itself still transfers: the 8-needle and 64k diagonals have fallen to near zero ($+0.1$), yet those rows still deliver substantial off-diagonal gains (up to $+2.1$ and $+1.2$), so a subset that no longer rewards training on itself remains a productive source for a stalled target. Thus no single fixed observation stays uniformly useful throughout post-training, the mechanistic reason adaptive observation can outperform fixed-view optimization.

\subsection{Harness evolution mitigates agent backend failures}
\label{sec:analysis:failures}

\begin{figure}[t]
    \centering
    \includegraphics[width=\columnwidth]{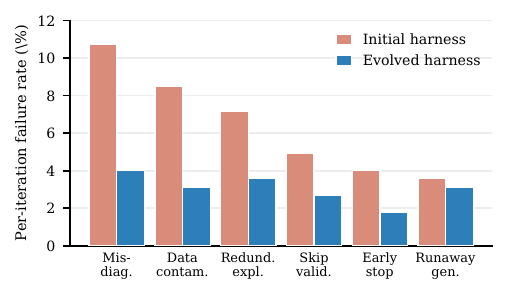}
    \caption{
    \textbf{Harness evolution reduces recurrent agent failures.}
    Per-iteration failure rate per defect type, before versus after evolution, averaged over backends; each bar is the fraction of all iterations whose dominant failure is that defect (see Table~\ref{tab:failure_modes}).
    }
    \label{fig:failures}
\end{figure}

Finally, we test whether harness evolution makes the post-training agent more reliable. Under matched tasks and per-iteration budgets, we label each failed iteration by its dominant defect, one of capability misdiagnosis, data contamination, redundant exploration, skipped validation, premature termination, and runaway generation, and compare the initial and evolved harnesses under both agent backends. Figure~\ref{fig:failures} shows that the first five defects, for which a harness update can encode a direct remedy, fall sharply after evolution, whereas runaway generation, largely a property of the backend decoder, falls only slightly. The reduction holds for both GLM-5.2 and Claude-Opus-4.8, attributing it to persistent harness-level updates rather than a stronger backend.

\section{Conclusion}
\label{sec:conclusion}

We identified \emph{observation adaptation} as a distinct, largely unexamined mechanism of harness evolution for post-training RSI. When the diagnostic granularity, difficulty, and tasks guiding the loop are frozen, coarse feedback hides heterogeneous gaps and early gains stall. Letting the harness adapt its observation, refining granularity, escalating difficulty, and probing related tasks, alongside the data it optimizes, re-exposes gaps the original feedback could not see. On Qwen3.5-35B-A3B, this sustains progress past the fixed-observation plateau, suggesting that sustaining RSI is as much about keeping feedback useful as about improving the artifacts being optimized.

\bibliography{custom}

\clearpage
\appendix

\setlength{\floatsep}{10pt plus 2pt minus 2pt}
\setlength{\textfloatsep}{7pt plus 2pt minus 2pt}
\setlength{\intextsep}{6pt plus 2pt minus 2pt}
\setlength{\dblfloatsep}{10pt plus 2pt minus 2pt}
\setlength{\dbltextfloatsep}{7pt plus 2pt minus 2pt}

\captionsetup[table]{skip=7pt}

\section{Limitations}
\label{sec:limitations}

\paragraph{Dependence on an adaptable observation.}
Adaptive observation is only useful when the loop actually has somewhere to move its view. Our three levers presuppose that a capability admits a finer diagnostic view that resolves the aggregate score into actionable failure modes (L1), a harder difficulty tier that is still measurable (L2), or a related task with a different surface form (L3). For a target that already saturates a single benchmark with no meaningful neighbor, no harder tier, and no informative sub-structure, our method reduces to the fixed-observation loop and offers no additional lever. Characterizing when a useful observation exists, and detecting its absence automatically rather than exhausting the preference order, is left to future work.

\paragraph{Coverage of capability directions.}
We validate adaptive observation in two separately trained long-context directions, reading comprehension and instruction following, using the same 35B mixture-of-experts base architecture. Whether the same plateau-breaking effect and lever preference order transfer to other domains, including mathematical and code reasoning, tool use, multilinguality, and multi-turn agentic behavior, and to substantially different model scales remains to be shown. Joint multi-direction training is outside the scope of our experiments; extending the loop across more directions and studying how their observations should be scheduled in a shared model is an important next step.

\paragraph{Reliance on benchmark-defined observations.}
Our observations are drawn from existing benchmarks and their difficulty tiers, so the reachable views are bounded by what public evaluations already expose. Capabilities for which no suitable benchmark, difficulty ladder, or fine-grained diagnostic exists cannot be observed adaptively without first constructing new evaluation instruments, which is outside the scope of this work.

\section{Implementation Details}
\label{sec:appendix}

\subsection{Observation controller}
\label{sec:appendix:observation}

The controller divides a run into observation phases. The active observation $O_r$ remains fixed within phase $r$, and its exponentially smoothed marginal-return history is reset whenever the controller commits $O_{r+1}$. The fixed anchor observation $O^{\mathrm{anc}}$ is evaluated in every phase. A candidate data increment is retained when it improves either the active or anchor score and does not reduce the anchor score beyond the regression tolerance; otherwise its checkpoint is discarded and the attempt is stored as negative evidence in $\mathcal{M}$.

At a plateau, let $\mathcal{F}_r$ be the concrete observation updates that are feasible under $O_r$ and have not already failed for the current gap. The controller selects $\ell_r=\operatorname{First}_{\textsc{L1}\prec\textsc{L2}\prec\textsc{L3}}(\mathcal{F}_r)$ and applies Eq.~\eqref{eq:lever-transitions}. Memory records the full attempted configuration, whether the refinement, tier, or related benchmark, rather than excluding an entire lever after one failed candidate.

For L2, the controller orders the benchmark's harder difficulty tiers $\delta_1\prec\cdots\prec\delta_J$ and chooses one with the help of a held-out open-source reference model $M_{\mathrm{ref}}$. We estimate $\widehat p_j=|\delta_j|^{-1}\sum_{x\in\delta_j}\mathbb{1}[M_{\mathrm{ref}}\text{ solves }x]$ and select
\begin{equation}
\label{eq:l2}
j^\star=\max\left\{j:p_{\mathrm{lo}}\le\widehat p_j\le p_{\mathrm{hi}}\right\},
\qquad
\delta^\star=\delta_{j^\star}.
\end{equation}
The band removes tiers that are already saturated and tiers that are effectively unsolvable. The chosen tier is added to the active observation, $\delta_{r+1}=\delta_r\cup\{\delta^\star\}$, so earlier tiers stay under observation rather than being replaced. For L3, the controller itself selects an untried benchmark $B^\star$ in the same capability direction but with a different task form and adds it to the cumulative active set, $\mathcal{B}_{r+1}=\mathcal{B}_r\cup\{B^\star\}$; earlier tasks remain active, and attempted related tasks are tracked individually in $\mathcal{M}$.

\paragraph{Default parameters.}
Unless otherwise noted, the controller uses the following settings. In Eq.~\eqref{eq:mr} we measure $\Delta d_t$ in units of $10$k retained increment examples, so the marginal return $\mathrm{MR}_t$ is expressed in score points per $10$k examples. We smooth it with weight $\alpha=0.3$ in Eq.~\eqref{eq:plateau} and declare a plateau once the smoothed return stays below $\epsilon=0.1$ throughout a trailing window of $w=3$ iterations. A candidate increment is retained when it improves the active or anchor score while reducing the anchor score by at most an anchor-regression tolerance of $\tau_{\mathrm{anc}}=0.5$ points. For the L2 tier selection of Eq.~\eqref{eq:l2}, the informative pass-rate band on the held-out reference model is $[p_{\mathrm{lo}},p_{\mathrm{hi}}]=[0.2,0.6]$, so the controller takes the hardest tier the reference model still solves at least $20\%$ but at most $60\%$ of the time.

\paragraph{Are the three levers mandatory?}
No. The ladder $\textsc{L1}\prec\textsc{L2}\prec\textsc{L3}$ fixes only the order in which levers are \emph{tried}: at each plateau the controller fires the first feasible lever and escalates to the next only when the current one yields no new gap (\S\ref{sec:method:adapt}), so a run uses as many or as few levers as its plateaus demand. To show that the levers are complementary rather than redundant, Table~\ref{tab:lever_ablation} restricts the controller to a single lever and compares it against the full ladder on four representative anchors. Each lever alone improves over the fixed-observation baseline but captures only part of the gain, and no single lever is best on any anchor; the two data-producing levers lead on the benchmarks they natively target: L2 drives the difficulty-scalable MRCR~v2 upto256k and LongGenBench, while L3 drives LongBench~v2 and LIFBench, where lateral tasks matter most. Only the full L1+L2+L3 ladder is best on every anchor, confirming that the gains come from coordinating the levers rather than from any single one.

\begin{table*}[t]
\centering
\caption{
\textbf{Lever ablation.} Restricting the controller to a single lever beats the fixed-observation baseline but recovers only part of the AdaObs gain, and no single lever is best on any anchor. FixedObs fires no lever; L1+L2+L3 is the AdaObs setting of Table~\ref{tab:main}. The two data-producing levers lead on the benchmarks they natively target: L2 on the difficulty-scalable MRCR~v2 upto256k and LongGenBench, and L3 on LongBench~v2 and LIFBench, where lateral tasks matter most.
}
\label{tab:lever_ablation}
\small
\setlength{\tabcolsep}{10pt}
\begin{tabular}{lccccc}
\toprule
\textbf{Configuration} & \makecell{LongBench\\v2} & \makecell{MRCR v2\\upto256k} & LIFBench & \makecell{LongGen\\Bench} & \textbf{Avg.} \\
\midrule
FixedObs   & 55.00 & 87.20 & 89.76 & 71.26 & 75.81 \\
\quad L1 only         & 55.74 & 90.33 & 90.42 & 73.10 & 77.40 \\
\quad L2 only         & 56.18 & 92.64 & 91.07 & 78.42 & 79.58 \\
\quad L3 only         & 57.21 & 88.76 & 92.53 & 74.05 & 78.14 \\
\rowcolor{oursrow}
L1+L2+L3 (AdaObs)              & \textbf{58.12} & \textbf{95.57} & \textbf{93.61} & \textbf{84.19} & \textbf{82.87} \\
\bottomrule
\end{tabular}
\end{table*}

\subsection{Harness implementation}
\label{sec:appendix:harness}

Let
\begin{equation}
H_k=\left(P_k,\,S_k,\,T_k,\,\mu_k,\,\pi_k^{\mathrm{obs}}\right)
\end{equation}
denote harness version $k$, where $P_k$ collects the \textsc{Analyze}, \textsc{Plan}, and \textsc{Produce} prompts, $S_k$ the agent's skills, $T_k$ its tool interfaces, $\mu_k$ its memory policy (what it writes to and retrieves from experiment memory $\mathcal{M}$), and $\pi_k^{\mathrm{obs}}$ the observation-controller policy. These prompts, skills, tools, memory policy, and controller policy are the only editable surfaces. The evaluator, anchor observation, contamination checks, artifact-provenance rules, the stored contents of $\mathcal{M}$, and the training implementation remain outside $H_k$.

At an update checkpoint, the system clusters recurring failures in experiment memory and proposes bounded candidates $\widetilde H_k^{(j)}=\Delta_j(H_k)$. Each candidate changes one editable surface and carries an audit record containing its target failure, intended effect, and regression risks. It is replayed on a repair set $\mathcal{R}_{\mathrm{fix}}$ containing the target failure and a regression buffer $\mathcal{R}_{\mathrm{reg}}$ containing successful and recent cross-observation decisions. Replay presents the original observation payload and scores the new decision against the original fixed contract and verifier.

Let $Q(H;\mathcal{R})$ be the fraction of replayed decisions that satisfy their contract and verifier. A candidate is accepted only if
\begin{equation}
\begin{aligned}
Q(\widetilde H_k^{(j)};\mathcal{R}_{\mathrm{fix}})
-Q(H_k;\mathcal{R}_{\mathrm{fix}})&\ge\eta,\\
Q(\widetilde H_k^{(j)};\mathcal{R}_{\mathrm{reg}})
-Q(H_k;\mathcal{R}_{\mathrm{reg}})&\ge-\varepsilon_H,
\end{aligned}
\label{eq:harness-gate}
\end{equation}
and passes static interface, provenance, and prompt-budget checks. The highest-scoring accepted candidate is committed as $H_{k+1}$; if none passes, the active harness remains $H_k$. By default we set the acceptance thresholds in Eq.~\eqref{eq:harness-gate} to $\eta=0.1$ and $\varepsilon_H=0.02$: a candidate must raise the fraction of repair-set decisions that satisfy their contract and verifier by at least $0.1$ while regressing the buffer by no more than $0.02$.

\paragraph{Control policy.}
Algorithm~\ref{alg:adaobs} states the adaptive-observation control policy referenced in \S\ref{sec:method:adapt}.

\begin{algorithm}[t]
\caption{Adaptive-Observation Post-Training Loop (one capability direction)}
\label{alg:adaobs}
\begin{algorithmic}[1]
\REQUIRE Base model $M_0$; anchor observation $O^{\mathrm{anc}}$; initial active observation $O$; base data $D_0$; plateau parameters $(\epsilon,w)$; reference model $M_{\mathrm{ref}}$; harness $H$
\ENSURE Retained increments $\Delta^\star$ and evolved harness $H$
\STATE $M\leftarrow M_0$; $\mathcal{M}\leftarrow\emptyset$; $\Delta^\star\leftarrow\emptyset$
\WHILE{compute budget remains}
    \STATE $\text{gap}\leftarrow\textsc{Analyze}(M;O,H,\mathcal{M})$
    \STATE $\iota\leftarrow\textsc{Plan}(\text{gap};H,\mathcal{M})$ \COMMENT{one falsifiable intervention}
    \STATE $\Delta_t\leftarrow\textsc{Validate}(\textsc{Produce}(\iota;H))$ \COMMENT{quality, provenance, contamination}
    \STATE $\widetilde M\leftarrow\textsc{Train}(M_0,D_0\cup\Delta^\star\cup\Delta_t)$
    \STATE $(\widetilde S^{\mathrm{act}},\widetilde S^{\mathrm{anc}})\leftarrow\textsc{Evaluate}(\widetilde M;O,O^{\mathrm{anc}})$
    \STATE update $\widehat{\mathrm{MR}}$ within the current observation phase
    \IF{active or anchor score improves, and anchor regression is within tolerance}
        \STATE $M\leftarrow\widetilde M$; $\Delta^\star\leftarrow\Delta^\star\cup\Delta_t$; append gain to $\mathcal{M}$
    \ELSE
        \STATE discard $\widetilde M$; append negative result to $\mathcal{M}$
    \ENDIF
    \IF{$\widehat{\mathrm{MR}}<\epsilon$ throughout the last $w$ iterations}
        \STATE $\ell\leftarrow\operatorname{First}_{\textsc{L1}\prec\textsc{L2}\prec\textsc{L3}}\{\ell:\operatorname{Feasible}(\ell;O,\mathcal{M})\}$
        \IF{no feasible lever remains}
            \STATE \textbf{break} \COMMENT{observation ladder exhausted}
        \ENDIF
        \STATE $O\leftarrow T_\ell(O)$ via Eq.~\eqref{eq:lever-transitions}; reset $\widehat{\mathrm{MR}}$
    \ENDIF
    \IF{harness-update checkpoint and a recurring editable failure exists}
        \STATE $H\leftarrow\textsc{EvolveHarness}(H,\mathcal{M})$ \COMMENT{mine, propose, replay, commit}
    \ENDIF
\ENDWHILE
\RETURN $(\Delta^\star,H)$
\end{algorithmic}
\end{algorithm}

\paragraph{Stage contracts.}
Every stage communicates through a fixed JSON contract: \textsc{Analyze}, \textsc{Plan}, \textsc{Produce}, \textsc{Validate}, \textsc{Train}, and \textsc{Evaluate} each declare a typed input and output schema that the execution layer validates at the stage boundary, so a malformed or out-of-contract decision is rejected before it reaches the cluster. Artifact-provenance gates attach the originating iteration, source data, and generation parameters to every increment, and the experiment memory $\mathcal{M}$ records each decision under a schema capturing its targeted gap, the intervention tried, the resulting active and anchor scores, and whether it was retained. Because the contract fixes exactly what each decision consumes and produces, a logged decision can be replayed verbatim during harness evolution (\S\ref{sec:method:harness}).

\subsection{Real-cluster infrastructure}
\label{sec:appendix:cluster}

The autonomous loop runs on a shared production cluster through a deterministic execution layer that sits between the agent's scientific decisions and the physical hardware. For each iteration the agent emits a typed plan, specifying the data to synthesize, the training configuration, and the evaluation suite, and the execution layer compiles it into a dependency graph of jobs, submits them, polls them to completion, and returns structured results to the next stage. The layer makes no scientific decisions; its sole responsibility is to make every iteration reproducible and recoverable so the loop can run unattended for days.

\paragraph{Job submission and node binding.}
Training and evaluation are dispatched as separate batch jobs to the cluster scheduler. Each full-parameter SFT job reserves 64 NVIDIA H200 GPUs across eight nodes (\S\ref{sec:appendix:training}); the launcher pins each rank to a fixed GPU and NUMA node, fixes the collective-communication topology and random seeds, and records the resolved host set so that a resubmission reproduces the same placement. Evaluation jobs are smaller and elastic: the layer fans the active and anchor benchmarks out across the remaining capacity and gathers their scores once all shards return.

\paragraph{Scheduler polling and checkpoint integrity.}
A supervisor process polls scheduler state at a fixed cadence and distinguishes transient faults (preemption, node drain, network timeouts), which are retried, from genuine job failures, which are surfaced to the agent as negative evidence. Before a checkpoint may seed the next iteration it passes an integrity gate: a shard-count and tensor-shape check against the training configuration, a checksum over the saved weights, and a short smoke-decode confirming that the checkpoint loads and emits well-formed output. A checkpoint failing any check is quarantined and its job resubmitted, preventing a silently corrupted artifact from propagating down the loop.

\paragraph{Persistent scheduling and recovery.}
All loop state, comprising the experiment memory $\mathcal{M}$, the retained increment set $\Delta^\star$, the active and anchor observations, and the status of every in-flight job, is written to a transactional store after each stage transition. On preemption or node failure the supervisor resumes from the last committed stage rather than restarting the iteration, and idempotent job keys ensure that a resubmitted job neither duplicates synthesized data nor double-counts an evaluation. This lets a run tolerate ordinary hardware churn with no human intervention.

\paragraph{Credential handling and run isolation.}
Model weights, agent-backend API keys, and dataset access tokens are injected into each job from a secrets manager at launch and are never written to logs, checkpoints, or the experiment memory. Synthesized data, trajectories, and checkpoints are stored under per-run namespaces, so the up to eight concurrent runs remain fully isolated from one another.

\paragraph{Contamination safeguards and configuration freezing.}
Every synthesized or re-rolled increment passes the contamination gate of \S\ref{sec:appendix:observation}, which runs $n$-gram and embedding-level overlap checks against all evaluation benchmarks, before it may enter training, so items resembling held-out test instances are removed at the source. The evaluator, anchor observations, decoding parameters, and validation gates are frozen at the start of a run and versioned alongside the harness; neither the agent nor the execution layer can alter the measurement apparatus mid-run, and any intended change produces a new, separately logged configuration rather than mutating the active one.

\subsection{Evaluation setup}
\label{sec:appendix:evaluation}

\paragraph{Benchmarks and protocols.}
We describe each observation benchmark and the size of the split we score below; exact subset definitions, per-benchmark difficulty tiers such as the MRCR~v2 upto256k needle-count tiers (2/4/8-needle), and the IFBench constraint-subset split are fixed across all runs. To control sampling variance, every reported score is an average over multiple independent evaluation passes (avg@$n$): we use avg@4 for LongBench~v2 and avg@3 for all other benchmarks in Table~\ref{tab:main} (MRCR~v2 upto256k, AA-LCR, SummHay, AdvancedIF, IFBench, LIFBench, and LongGenBench). On PostTrainBench (Table~\ref{tab:ptb}) every sub-benchmark is reported as avg@4.

\paragraph{Decoding and evaluation parameters.}
Unless stated otherwise, all models are evaluated with thinking (long-CoT) enabled, and we use a single fixed decoding configuration across every benchmark, iteration, and run so that score differences reflect the training data rather than the sampling setup. We decode with temperature $0.8$ and a maximum generation length of $64$K tokens on every benchmark. For the two longest-context benchmarks we extend the context window via RoPE scaling: LongBench~v2 uses a RoPE scaling factor of $4$ with a $1$M-token session length, and MRCR~v2 upto256k uses a RoPE scaling factor of $2$ with a $512$K-token session length; the remaining benchmarks run at the model's native $256$K-token context length. Outputs that exhaust the generation budget without producing a parseable answer are counted as failures and contribute to the over-length response ratio analyzed in \S\ref{sec:analysis:cot}. For avg@$n$ benchmarks we draw $n$ independent samples under these settings and average the per-sample scores. The CoT-supervision analysis in \S\ref{sec:analysis:cot} varies the training mixture and evaluation regime as specified there while keeping all other decoding settings fixed.

\smallskip\noindent\textit{Reading comprehension.}
\begin{itemize}[leftmargin=1.2em,itemsep=1pt,topsep=1pt]
    \item \textbf{LongBench~v2}~\citep{bai2025longbench}: 503 challenging multiple-choice questions over contexts from 8k to 2M words, spanning six task categories: single- and multi-document QA, long in-context learning, dialogue, code-repository, and structured-data understanding. Scored by multiple-choice accuracy.
    \item \textbf{MRCR~v2 upto256k}~\citep{deepmind2026mrcr}: multi-round co-reference resolution, where the model must retrieve a specific earlier response among several near-duplicate ``needles'' in a long synthetic dialogue, under 2/4/8-needle settings. We score the up-to-256k context range, $2{,}167$ questions in total; for the observation-refinement analysis (Figure~\ref{fig:difficulty}) we split it by needle count into 2-needle ($723$ questions), 4-needle ($722$ questions), and 8-needle ($722$ questions) tiers of increasing difficulty, and the main table reports their question-count-weighted combination.
    \item \textbf{AA-LCR}~\citep{aa2025lcr}: 100 hard reasoning questions, each requiring synthesis across multiple real-world documents, roughly 100k tokens per question, drawn from academic papers, financial filings, legal, and government texts.
    \item \textbf{SummHay}~\citep{laban2024summary}: query-focused summarization over synthesized ``haystacks'' of $\sim$100 documents, around 100k tokens, in which specific insights repeat; systems must produce a summary citing the relevant documents. We evaluate $92$ query-haystack instances, scored on coverage and citation quality.
\end{itemize}

\smallskip\noindent\textit{Instruction following.}
\begin{itemize}[leftmargin=1.2em,itemsep=1pt,topsep=1pt]
    \item \textbf{AdvancedIF}~\citep{he2026advancedif}: an advanced instruction-following benchmark of over 1{,}600 expert-written prompts with expert-curated rubrics, each prompt carrying multiple composed constraints spanning format, style, structure, length, and negative constraints; scored against its rubric.
    \item \textbf{IFBench}~\citep{pyatkin2026generalizing}: 58 new, diverse, verifiable constraints applied to 300 held-out prompts sourced from WildChat, measuring precise instruction-following generalization via per-constraint verification functions.
    \item \textbf{LIFBench}~\citep{wu2025lifbench}: $2{,}802$ instructions across three long-context scenarios, List, MultiDoc, and OneDoc, and eleven tasks, evaluating instruction-following performance and stability in long contexts. Each scenario is instantiated at six context-length intervals from $4$k to $128$k tokens, which provides the input-length difficulty axis used in the observation-refinement analysis.
    \item \textbf{LongGenBench}~\citep{liu2024longgenbench}: constrained long-form generation over four scenarios and three instruction types at two target generation lengths (16K and 32K tokens), scoring whether instructions are satisfied throughout a very long output.
\end{itemize}

\subsection{Training setup and compute}
\label{sec:appendix:training}

\paragraph{SFT hyperparameters.}
Every experiment runs full-parameter SFT from Qwen3.5-35B-A3B-base~\citep{qwen35}. Table~\ref{tab:sft} lists the default configuration all runs start from; the training data, the base plus accumulated increments, differs across runs, and the fixed-observation loop may additionally retune these recipe knobs when it has no new observation to act on (Table~\ref{tab:case_trace}). We use a maximum sample length and pack length of $256\times1024$ tokens, so sequences are packed to $256$K tokens, a global batch size of $16$, and the AdamW optimizer with a peak learning rate of $2\times10^{-5}$ decayed cosine-wise to a minimum of $1\times10^{-6}$, a warmup ratio of $0.1$, and weight decay of $0.05$. We enable full activation recomputation at recompute ratio $1.0$ and use square-root loss reduction, \texttt{loss\_reduction = "square"}, to balance the contribution of packed sequences of differing lengths.

\begin{table}[H]
\centering
\caption{
\textbf{Default SFT training configuration} that all runs start from; the training data differs across runs, and the fixed-observation loop may retune these recipe knobs when it has no new observation to act on (Table~\ref{tab:case_trace}).
}
\label{tab:sft}
\small
\setlength{\tabcolsep}{6pt}
\begin{tabular}{ll}
\toprule
\textbf{Hyperparameter} & \textbf{Value} \\
\midrule
Base model            & Qwen3.5-35B-A3B-base \\
Tuning                & full-parameter SFT (no RL) \\
Sample max length     & $256\times1024$ tokens \\
Pack max length       & $256\times1024$ tokens \\
Global batch size     & $16$ \\
Optimizer             & AdamW \\
Peak learning rate    & $2\times10^{-5}$ \\
Min learning rate     & $1\times10^{-6}$ \\
LR schedule           & cosine decay \\
Warmup ratio          & $0.1$ \\
Weight decay          & $0.05$ \\
Recompute ratio       & $1.0$ \\
Loss reduction        & square \\
Hardware              & $64\times$ NVIDIA H200 \\
\bottomrule
\end{tabular}
\end{table}

\paragraph{Compute.}
Each experiment is trained on 64 NVIDIA H200 GPUs, taking $\sim$12h per iteration, $\sim$2 iterations per run per day, with up to $\sim$8 concurrent runs.

\paragraph{Increment-data budget.}
The two routes are matched in everything but the observation. Both start from the same 450k base and the same recipe, run under an identical per-run cap on the number of loop iterations, and let the same agent decide autonomously what data to synthesize or download, how much, and when to stop; the only difference is that FixedObs always analyzes the anchor-only aggregate, whereas AdaObs may adapt its active observation at a plateau. The data selection is therefore the agent's own choice in both routes rather than a hand-set budget, so the only quantity that differs across runs is the retained increment data each best checkpoint ends up trained on. Table~\ref{tab:increment_budget} reports this budget per anchor as training tokens, for the best FixedObs and AdaObs checkpoints of Table~\ref{tab:main}. On every benchmark the two routes stay within the same order of magnitude, with AdaObs larger on some anchors and smaller on others and comparable in total, yet AdaObs scores higher on all eight. This rules out a larger data budget or more iterations as the source of its gains and isolates the observation itself as the active ingredient.

\begin{table*}[t]
\centering
\caption{
\textbf{Increment-data budget (training tokens) of each best checkpoint, FixedObs vs.\ AdaObs.} Tokens of the retained increment data beyond the shared 450k base, for the best checkpoints of Table~\ref{tab:main}. Both routes stay within the same order of magnitude on every anchor, so AdaObs's higher scores (Table~\ref{tab:main}) come from the observation, not a larger data budget.
}
\label{tab:increment_budget}
\footnotesize
\setlength{\tabcolsep}{5pt}
\begin{tabular}{lcccccccc}
\toprule
\multirow{2}{*}{\textbf{Model}} & \multicolumn{4}{c}{\textbf{Reading Comprehension}} & \multicolumn{4}{c}{\textbf{Instruction Following}} \\
\cmidrule(lr){2-5} \cmidrule(lr){6-9}
 & \makecell{LongBench\\v2} & \makecell{MRCR v2\\upto256k} & AA-LCR & SummHay & \makecell{Advanced\\IF} & IFBench & LIFBench & \makecell{LongGen\\Bench} \\
\midrule
FixedObs       & 0.28B & 0.19B & 0.21B & 0.26B & 0.14B & 0.20B & 0.23B & 0.15B \\
\rowcolor{oursrow}
\textbf{AdaObs (ours)} & 0.31B & 0.15B & 0.25B & 0.22B & 0.16B & 0.17B & 0.20B & 0.17B \\
\bottomrule
\end{tabular}
\end{table*}

\subsection{Score variability}
\label{sec:appendix:controls}

To quantify evaluation noise, Table~\ref{tab:main_std} and Table~\ref{tab:ptb_std} report mean\,$\pm$\,standard deviation over the independent evaluation passes behind Table~\ref{tab:main} and Table~\ref{tab:ptb}, respectively. For the main results, the std is taken over the $n$ passes of each avg@$n$ benchmark. On PostTrainBench each cell is the mean over the four post-trained student models, namely Qwen3-1.7B and Qwen3-4B~\citep{qwen35}, SmolLM3-3B~\citep{smollm3}, and Gemma3-4B~\citep{team2025gemma}, each scored with avg@4; the reported std is the evaluation sampling noise on this mean, computed over the independent avg@4 passes and pooled across the four students. We deliberately do \emph{not} report the raw spread \emph{across} the four student scales as an error bar: students differ in size by more than $2\times$, so that spread reflects an expected and systematic capacity gap between models rather than measurement noise, and folding it into the error bar would conflate two distinct effects. The deviations are small relative to the gaps between methods, so the ordering in Table~\ref{tab:main} and Table~\ref{tab:ptb} is not an artifact of sampling noise.

\begin{table*}[tp]
\centering
\caption{\textbf{Score variability on Table~\ref{tab:main} (mean\,$\pm$\,std).}}
\label{tab:main_std}
\small
\setlength{\tabcolsep}{4pt}
\resizebox{\textwidth}{!}{
\begin{tabular}{lcccccccc}
\toprule
\multirow{2}{*}{\textbf{Model}} & \multicolumn{4}{c}{\textbf{Reading Comprehension}} & \multicolumn{4}{c}{\textbf{Instruction Following}} \\
\cmidrule(lr){2-5} \cmidrule(lr){6-9}
 & \makecell{LongBench\\v2} & \makecell{MRCR v2\\upto256k} & AA-LCR & SummHay & \makecell{Advanced\\IF} & IFBench & LIFBench & \makecell{LongGen\\Bench} \\
\midrule
FixedObs & \ms{55.00}{0.62} & \ms{87.20}{0.58} & \ms{66.00}{0.81} & \ms{38.69}{0.55} & \ms{63.12}{0.82} & \ms{60.51}{0.74} & \ms{89.76}{0.52} & \ms{71.26}{0.71} \\
\rowcolor{oursrow}
AdaObs (ours)      & \ms{58.12}{0.48} & \ms{95.57}{0.50} & \ms{72.00}{0.69} & \ms{50.07}{0.47} & \ms{68.86}{0.70} & \ms{66.88}{0.63} & \ms{93.61}{0.44} & \ms{84.19}{0.60} \\
\bottomrule
\end{tabular}
}
\end{table*}

\begin{table*}[tp]
\centering
\caption{\textbf{Score variability on Table~\ref{tab:ptb} (mean\,$\pm$\,std).}}
\label{tab:ptb_std}
\small
\setlength{\tabcolsep}{5pt}
\resizebox{\textwidth}{!}{
\begin{tabular}{llcccccccc}
\toprule
\textbf{Harness} & \textbf{Agent backend} & \makecell{AIME\\2025} & GSM8K & GPQA & HumanEval & BFCL & \makecell{ArenaHard\\-Writing} & \makecell{HealthBench\\-Easy} & \textbf{Avg.} \\
\midrule
Default harness & GLM-5.2          & \ms{5.2}{0.6} & \ms{61.8}{1.4} & \ms{25.6}{1.0} & \ms{46.5}{1.6} & \ms{46.1}{2.0} & \ms{8.4}{0.7} & \ms{18.2}{0.9} & \ms{30.3}{1.1} \\
\rowcolor{oursrow}
AdaObs (ours)   & GLM-5.2          & \ms{6.7}{0.7} & \ms{69.3}{1.3} & \ms{28.8}{1.1} & \ms{54.6}{1.5} & \ms{70.7}{1.9} & \ms{10.7}{0.8} & \ms{19.4}{0.9} & \ms{37.2}{1.2} \\
\midrule
Default harness & Claude-Opus-4.8  & \ms{7.9}{0.6} & \ms{67.8}{1.3} & \ms{30.2}{1.0} & \ms{54.6}{1.5} & \ms{55.1}{2.0} & \ms{11.7}{0.8} & \ms{22.6}{0.9} & \ms{35.7}{1.1} \\
\rowcolor{oursrow}
AdaObs (ours)   & Claude-Opus-4.8  & \ms{9.7}{0.7} & \ms{73.8}{1.2} & \ms{33.2}{1.1} & \ms{61.6}{1.4} & \ms{78.3}{1.8} & \ms{15.1}{0.9} & \ms{24.4}{1.0} & \ms{42.3}{1.2} \\
\bottomrule
\end{tabular}
}
\end{table*}

\subsection{Observation-refinement breakdowns}
\label{sec:appendix:difficulty}

Tables~\ref{tab:mrcr_needle}, \ref{tab:longgen_len}, \ref{tab:longbench_related}, and~\ref{tab:lifbench_related} give the exact per-stage scores plotted in Figure~\ref{fig:difficulty}, covering the two data-producing levers. Each panel traces a single lever stage on its \emph{own} anchor run, with the horizontal axis re-zeroed to that lever's entry and measuring iterations \emph{within} the stage, as the lever adds its subsets or related tasks one at a time. This within-stage clock is finer than the macro per-iteration axis of the full runs in Figure~\ref{fig:main-exp}, where a lever appears as a single firing point followed by a few fused iterations; the panels are therefore not sub-windows of Figure~\ref{fig:main-exp}, and panels (a,b) are MRCR~v2 upto256k and LongGenBench anchors that do not appear in Figure~\ref{fig:main-exp} at all. The \emph{difficulty} lever (L2) refines a benchmark into harder within-benchmark subsets: MRCR~v2 upto256k by needle count (2/4/8-needle) and LongGenBench by target output length (16k/32k). In each L2 table a row is a subset of that benchmark and a column is an observation stage: the first column observes only the first subset, and each later column \emph{adds} the next, harder subset to the observation; hence the ``$+$'' in the column headers, where ``$+$4-needle'' means the observation now also includes the 4-needle subset. A cell is the best score the model reaches on that row's subset under that column's observation, so every cell is filled, and reading across a row shows the effect of interest: each subset already under observation keeps climbing as later subsets are added, so that the 2-needle subset rises $94.00\to96.40\to98.57$ as the 4- and 8-needle subsets enter, and the earliest-observed subset ends strongest even though each step nominally targets the newly added subset.

\begin{table}[t]
\centering
\caption{\textbf{MRCR~v2 upto256k by needle count across observation stages.}}
\label{tab:mrcr_needle}
\small
\setlength{\tabcolsep}{5pt}
\begin{tabular}{lccc}
\toprule
\textbf{Needle count} & 2-needle & $+$4-needle & $+$8-needle \\
\midrule
2-needle & 94.00 & 96.40 & 98.57 \\
4-needle & 87.20 & 91.50 & 95.57 \\
8-needle & 80.40 & 86.80 & 92.57 \\
\bottomrule
\end{tabular}
\end{table}

\begin{table}[t]
\centering
\caption{\textbf{LongGenBench by output length across observation stages.}}
\label{tab:longgen_len}
\small
\setlength{\tabcolsep}{5pt}
\begin{tabular}{lcc}
\toprule
\textbf{Output length} & 16k & $+$32k \\
\midrule
16k & 84.03 & 94.72 \\
32k & 58.48 & 73.66 \\
\bottomrule
\end{tabular}
\end{table}

The \emph{related-task} lever (L3) instead expands the benchmark set within the same capability direction: LongBench~v2 as AA-LCR then MRCR~v2 upto256k enter (Table~\ref{tab:longbench_related}), and LIFBench as IFBench then AdvancedIF enter (Table~\ref{tab:lifbench_related}). Here each column adds one \emph{related benchmark} to the observation, and reading across the primary benchmark's row shows it stepping up again (LongBench~v2 $55.00\to56.63\to58.12$; LIFBench $89.76\to92.14\to93.61$) each time a same-direction task enters, renewed gains that come from changing \emph{which task} is observed rather than from harder examples of the original benchmark.

In each panel of Figure~\ref{fig:difficulty} the grey FixedObs curve warms up quickly and then keeps only a slow residual rise, far below AdaObs, ending exactly on the FixedObs value it is compared against: the \emph{subset-level} FixedObs in (a,b) ($94.00$ for 2-needle and $84.03$ for 16k, the first column of Tables~\ref{tab:mrcr_needle} and~\ref{tab:longgen_len}) and the \emph{benchmark-level} FixedObs of Table~\ref{tab:main} in (c,d) ($55.00$ for LongBench~v2 and $89.76$ for LIFBench).

\begin{table}[t]
\centering
\caption{\textbf{LongBench~v2 as related RC tasks enter the observation (L3).} Rows are benchmarks; columns are observation stages that \emph{add} one related RC benchmark at a time, all measured on a \emph{single} LongBench-anchored run. The anchor row (LongBench~v2) spans its fixed-observation plateau to its Table~\ref{tab:main} endpoint; the co-observed rows rise through transfer but stay below the dedicated endpoints their own anchor runs reach in Table~\ref{tab:main}.}
\label{tab:longbench_related}
\small
\setlength{\tabcolsep}{3pt}
\begin{tabular}{lccc}
\toprule
\textbf{Benchmark} & \makecell{Long-\\Bench v2} & \makecell{$+$AA-\\LCR} & \makecell{$+$MRCR v2\\upto256k} \\
\midrule
LongBench v2 & 55.00 & 56.63 & 58.12 \\
AA-LCR       & 57.83 & 61.46 & 64.91 \\
MRCR v2 upto256k & 79.64 & 84.27 & 88.93 \\
\bottomrule
\end{tabular}
\end{table}

\begin{table}[t]
\centering
\caption{\textbf{LIFBench as related IF tasks enter the observation (L3).} Rows are benchmarks; columns are observation stages that \emph{add} one related IF benchmark at a time, all measured on a \emph{single} LIFBench-anchored run. The anchor row (LIFBench) spans its fixed-observation plateau to its Table~\ref{tab:main} endpoint; the co-observed rows rise through transfer but stay below the dedicated endpoints their own anchor runs reach in Table~\ref{tab:main}.}
\label{tab:lifbench_related}
\small
\setlength{\tabcolsep}{5pt}
\begin{tabular}{lccc}
\toprule
\textbf{Benchmark} & LIFBench & $+$IFBench & $+$AdvIF \\
\midrule
LIFBench   & 89.76 & 92.14 & 93.61 \\
IFBench    & 52.87 & 56.63 & 59.88 \\
AdvancedIF & 54.91 & 58.26 & 61.37 \\
\bottomrule
\end{tabular}
\end{table}

\paragraph{Failure-mode breakdown.}
Table~\ref{tab:failure_modes} gives the per-backend values summarized in Figure~\ref{fig:failures}: for each defect type, the fraction of \emph{all} iterations whose dominant failure is that defect, under the initial and evolved harness, for both the GLM-5.2 and Claude-Opus-4.8 agent backends, together with the backend average. Each iteration is labeled by its single dominant defect, so a column sums to that configuration's overall per-iteration failure rate (last row), not to $100\%$; successful iterations and any residual defects outside these six categories make up the remainder. The two backends show the same pattern: the harness-fixable defects fall substantially while runaway generation falls only slightly, so the reduction tracks the harness rather than the backend.

\begin{table}[H]
\centering
\caption{\textbf{Per-iteration failure rate (\%) by dominant defect type}, for the initial (I) and evolved (E) harness under each agent backend, with the backend average plotted in Figure~\ref{fig:failures}. Each entry is the share of \emph{all} iterations of that configuration whose dominant failure is the given defect, snapped to the nearest feasible $n/224$ and reported to two decimal places.}
\label{tab:failure_modes}
\small
\setlength{\tabcolsep}{3pt}
\resizebox{\columnwidth}{!}{%
\begin{tabular}{lcccccc}
\toprule
 & \multicolumn{2}{c}{GLM-5.2} & \multicolumn{2}{c}{Opus-4.8} & \multicolumn{2}{c}{Avg.} \\
\cmidrule(lr){2-3}\cmidrule(lr){4-5}\cmidrule(lr){6-7}
\textbf{Defect type} & I & E & I & E & I & E \\
\midrule
Capability misdiagnosis & 11.16 & 4.46 &  9.82 & 3.57 & 10.71 & 4.02 \\
Data contamination      &  8.93 & 3.57 &  8.04 & 2.68 &  8.48 & 3.13 \\
Redundant exploration   &  7.59 & 4.02 &  6.70 & 3.13 &  7.14 & 3.57 \\
Skipped validation      &  5.36 & 3.13 &  4.46 & 1.79 &  4.91 & 2.68 \\
Premature termination   &  4.46 & 2.68 &  3.57 & 1.34 &  4.02 & 1.79 \\
Runaway generation      &  4.02 & 3.57 &  3.13 & 2.68 &  3.57 & 3.13 \\
\midrule
Overall failure rate    & 41.52 & 21.43 & 35.72 & 15.19 & 38.83 & 18.32 \\
\bottomrule
\end{tabular}
}
\end{table}

\subsection{Why the fixed observation plateaus: a LongBench~v2 case study}
\label{sec:appendix:plateau}

To make concrete why a frozen observation stalls, we trace one anchor, LongBench~v2, under the fixed-observation route and contrast it with what the adaptive controller sees at the \emph{same} plateau. Both routes start from the same base model and the same initial increment, and both optimize the single aggregate LongBench~v2 accuracy as the target; the only difference is what each is allowed to \emph{observe} between iterations.

\paragraph{The fixed-observation route exhausts its knobs.}
Under one aggregate score the loop can turn only two kinds of knob: re-balance the mixture of already-downloaded public reading-comprehension data, or retune the training recipe (learning rate, epochs, schedule, batch size). Table~\ref{tab:case_trace} logs seven iterations of exactly this: each knob moves the aggregate by at most a few tenths, and after the sixth iteration the smoothed marginal return stays below $\epsilon$ for the full trailing window, so a plateau is declared at $55.0$. Every remaining action is a variation on data it already has or a recipe it has already tuned.

\paragraph{The aggregate hides a context-length cliff.}
Why none of these knobs helps becomes obvious only once the score is resolved by input length (Table~\ref{tab:case_length}). The downloaded corpus is dominated by short inputs ($85\%$ of SFT items under $32$k tokens, only $3\%$ above $128$k), whereas LongBench~v2 places a quarter of its evaluation mass above $128$k. The short bucket is already near its ceiling ($68.0$) and flat, so reweighting toward ``long-document'' sources that are themselves short only nudges a saturated bucket; the genuinely long bucket sits at the floor ($36.8$) and does not move, because no training knob can supply long-context behavior the data never contained. The aggregate $55.0$ is a blend of the two, and this averaging is precisely what lets a localized failure read as ``no further gains available.'' The long bucket also carries a high over-length ratio, where the model exhausts the $64$k generation budget without a parseable answer (\S\ref{sec:appendix:evaluation})---another signal the single number suppresses.

\paragraph{The L1 lever exposes the gap the aggregate hid.}
At the same plateau the adaptive controller fires its cheapest lever, L1, which refines diagnostic granularity rather than proposing data: it decomposes LongBench~v2 by input-length bucket and over-length indicator, surfacing the heterogeneity in Table~\ref{tab:case_length}. \textsc{Analyze} emits a gap record whose \texttt{target} is the $>128$k bucket, whose \texttt{evidence} is the flat floor score and elevated over-length ratio there, and whose \texttt{hypothesis} is that the increment lacks genuinely long-context supervision. \textsc{Plan} turns this into a falsifiable intervention---add long-context SFT data above $128$k tokens rather than more short data, with the long-bucket score as \texttt{falsifier}---and \textsc{Produce} synthesizes and downloads data meeting that profile under the contamination gate. On the next checkpoint the long bucket climbs $36.8\to39.4$ while the short bucket barely moves, and the aggregate breaks the plateau to $55.7$, matching the L1-only ablation in Table~\ref{tab:lever_ablation}. The gain comes entirely from \emph{changing what was observed}. L1 only re-opens the gap; the controller then escalates to L2 (difficulty) and L3 (related tasks) on the same anchor, carrying LongBench~v2 the rest of the way to its $58.12$ endpoint in Table~\ref{tab:main}.

\begin{table}[t]
\centering
\caption{\textbf{Fixed-observation trace on LongBench~v2.} Under a single aggregate score the loop can only re-balance already-downloaded data or retune the recipe. Each knob moves the aggregate by at most a few tenths, and after the sixth iteration the smoothed marginal return stays below $\epsilon$ for the trailing window: a plateau is declared at $55.0$.}
\label{tab:case_trace}
\small
\setlength{\tabcolsep}{3pt}
\renewcommand{\arraystretch}{1.12}
\begin{tabularx}{\columnwidth}{@{}c >{\raggedright\arraybackslash}X r@{}}
\toprule
\textbf{It.} & \textbf{Adjustment under the aggregate observation} & \textbf{Score} \\
\midrule
1 & Initial increment, base SFT recipe & 53.4 \\
2 & Reweight mixture toward long-document QA sources & 54.1 \\
3 & Download more same-profile reading-comprehension data & 54.5 \\
4 & Lower learning rate ($2\mathrm{e}{-}5\!\to\!1\mathrm{e}{-}5$) & 54.7 \\
5 & Train one extra epoch ($2\!\to\!3$) & 54.9 \\
6 & Retune schedule: longer warmup, larger batch & 55.0 \\
7 & Re-balance mixture again toward long-document QA & 55.0 \\
\midrule
\multicolumn{2}{@{}l}{\emph{Plateau declared} ($\widehat{\mathrm{MR}}<\epsilon$)} & \textbf{55.0} \\
\bottomrule
\end{tabularx}
\end{table}

\begin{table}[t]
\centering
\caption{\textbf{What the aggregate hides on LongBench~v2.} Resolving the frozen score by input length exposes a context-length cliff. The downloaded public corpus is $85\%$ short while a quarter of the evaluation mass exceeds $128$k tokens, so the short bucket is already saturated and flat while the long bucket sits at the floor; the recipe and mixture knobs of Table~\ref{tab:case_trace} can only nudge the saturated bucket. Firing L1 (granularity) resolves the score into these buckets and targets the long one with genuinely long-context data, re-opening the plateau to $55.7$ (the L1-only ablation of Table~\ref{tab:lever_ablation}); the later L2 and L3 levers then carry the anchor to its $58.12$ endpoint in Table~\ref{tab:main}.}
\label{tab:case_length}
\small
\setlength{\tabcolsep}{5pt}
\begin{tabular}{lcccc}
\toprule
\textbf{Input length} & \makecell{Eval\\share} & \makecell{SFT-data\\share} & \makecell{FixedObs\\(plateau)} & \makecell{AdaObs\\(after L1)} \\
\midrule
$<32$k        & 35\% & 85\% & 68.0 & 68.2 \\
$32$--$128$k  & 40\% & 12\% & 55.0 & 55.0 \\
$>128$k       & 25\% &  3\% & 36.8 & \textbf{39.4} \\
\midrule
Aggregate     & 100\% & 100\% & \textbf{55.0} & \textbf{55.7} \\
\bottomrule
\end{tabular}
\end{table}

\end{document}